\documentclass[compress,times,preprint,review,10pt]{elsarticle}

\usepackage{amssymb}
\usepackage{wrapfig}
\usepackage{multirow}
\usepackage{hyperref}
\usepackage{amsmath}
\usepackage{booktabs}

\journal{Pattern Recognition}

\begin{document}

\begin{frontmatter}


\author[a]{Congqi Cao\corref{*}}
\ead{congqi.cao@nwpu.edu.cn}
\author[a]{Peiheng Han}
\author[a]{Yueran Zhang}
\author[a]{Yating Yu}
\author[b]{Qinyi Lv}
\author[b]{Lingtong Min}
\author[a]{Yanning Zhang}
\cortext[*]{Corresponding author}
\affiliation[a]{organization={School of Computer Science, Northwestern Polytechnical University}, country={China}}
\affiliation[b]{organization={School of Electronics and Information, Northwestern Polytechnical University}, country={China}}

\title{Task-Adapter++: Task-specific Adaptation with Order-aware Alignment for Few-shot Action Recognition}


\begin{abstract}
Large-scale pre-trained models have achieved remarkable success in language and image tasks, leading an increasing number of studies to explore the application of pre-trained image models, such as CLIP, in the domain of few-shot action recognition (FSAR). However, current methods generally suffer from several problems: 1) Direct fine-tuning often undermines the generalization capability of the pre-trained model; 2) The exploration of task-specific information is insufficient in the visual tasks; 3) The semantic order information is typically overlooked during text modeling; 4) Existing cross-modal alignment techniques ignore the temporal coupling of multimodal information. To address these, we propose Task-Adapter++, a parameter-efficient dual adaptation method for both image and text encoders. 
Specifically, to make full use of the variations across different few-shot learning tasks, we design a task-specific adaptation for the image encoder so that the most discriminative information can be well noticed during feature extraction. 
Furthermore, we leverage large language models (LLMs) to generate detailed sequential sub-action descriptions for each action class, and introduce semantic order adapters into the text encoder to effectively model the sequential relationships between these sub-actions. 
Finally, we develop an innovative fine-grained cross-modal alignment strategy that actively maps visual features to reside in the same temporal stage as semantic descriptions. Extensive experiments fully demonstrate the effectiveness and superiority of the proposed method, which achieves state-of-the-art performance on 5 benchmarks consistently. The code is open-sourced at \href{https://github.com/Jaulin-Bage/Task-Adapter-pp}{https://github.com/Jaulin-Bage/Task-Adapter-pp}.
\end{abstract}

\begin{graphicalabstract}
\end{graphicalabstract}

\begin{highlights}
\item We propose Task-Adapter++, a parameter-efficient dual adaptation framework for few-shot action recognition, which alleviates catastrophic forgetting and overfitting induced by full fine-tuning.
\item For the visual branch, we propose Task-Adapter to extract discriminative task-specific information during feature extraction.
\item For the semantic branch, we propose semantic order adaptation, which utilizes dynamic descriptions to model the sequential relationships among sub-actions, thereby enhancing the performance of stage-wise cross-modal matching.
\item Task-Adapter++ achieves state-of-the-art performance across five commonly-used few-shot action recognition benchmarks.
\end{highlights}


\end{frontmatter}


\begin{figure*}[h]
  \centering
  \includegraphics[width=1.0\textwidth]{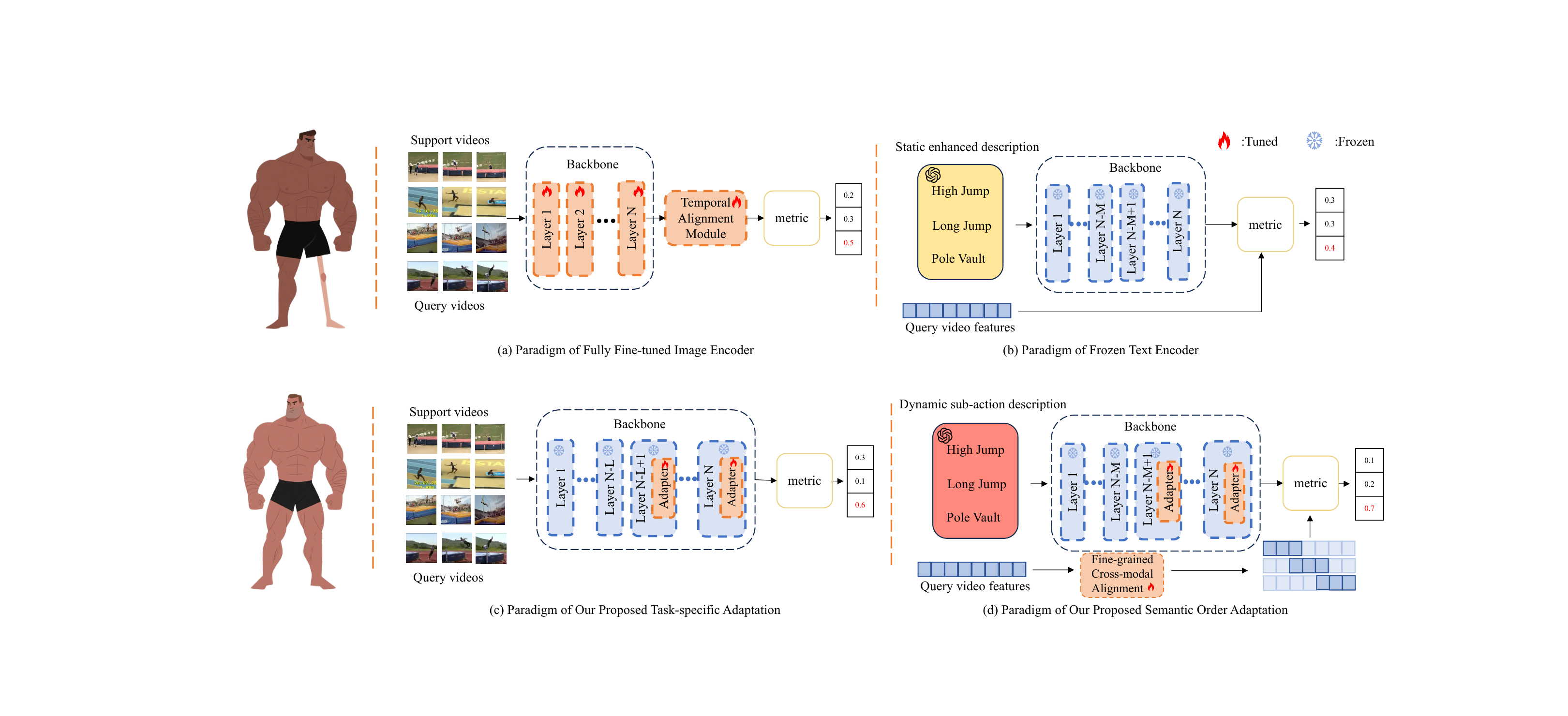}
  \caption{Comparison of different paradigms. (a) Existing methods fully fine-tune the feature extractor and combine it with temporal alignment module. (b) Existing methods extract semantic features from static enhanced descriptions and calculate the score by global matching. (c) For the image encoder, we freeze the backbone and apply task-specific adaptation to extract the most discriminative information. (d) For the text encoder, we use semantic order adapter to model sequential information within dynamic sub-actions and further calculate the score by stage with aligned visual features.}
  \label{fig:fine-tune}
\end{figure*}

    
\section{Introduction}
\label{section1}
With the advent of large-scale pre-trained models~\cite{radford2021learning,li2022blip}, researchers are now able to exploit the knowledge these models have acquired during pre-training and apply it to downstream tasks. Accordingly, there is an increasing focus on fine-tuning large pre-trained models for specific tasks~\cite{gao2023clip,zhou2023zegclip,yang2023aim,wortsman2022robust}.
In video understanding, existing works typically fine-tune pre-trained models on well-labeled video datasets~\cite{carreira2017quo,goyal2017something}, where the training and testing categories are identical. However, in realistic scenarios, collecting sufficient and well-labeled video samples is inherently challenging. As a result, few-shot learning has garnered increasing attention~\cite{peng2019few,chen2019closer,tian2020rethinking,ye2020few,cao2022learning}. 

A common practice in few-shot action recognition is to adopt a metric-based meta-learning strategy, where classification probabilities are computed based on the distance between query video features and support prototypes.
Prior works~\cite{cao2020few,perrett2021temporal,wang2022hybrid,li2022ta2n} achieve this by fine-tuning large image models, such as CLIP~\cite{radford2021learning}, on video datasets alongside independent temporal modules, as shown in Fig.~\ref{fig:fine-tune} (a). 
Beyond visual-centric approaches, recent studies have explored multimodal frameworks for few-shot learning by integrating semantic information. For instance, works like~\cite{wang2023few,xing2023multimodal} utilize the zero-shot classification ability of CLIP by video-text matching, significantly improving the accuracy of few-shot classification. Some works~\cite{yu2024building,Li_Liu_Wang_Yu_2025,DU2025111283} additionally employ large language models (LLMs) or vision-language models (VLMs) to expand label information, enriching the attribute information of actions or objects, as shown in Fig.~\ref{fig:fine-tune} (b). 

Although previous studies have made progress, several limitations remain. First, full fine-tuning erodes the model’s inherent prior knowledge, thereby degrading its ability to generalize to unseen tasks. Second, the importance of incorporating task-specific adaptation into spatio-temporal feature extraction is often overlooked, which leads to the neglect of the most discriminative information in each task. Third, the enhanced label information is typically treated as a static description, ignoring the different stages of action occurrence. Due to the absence of fine-grained temporal descriptions and semantic order modeling, it is difficult to fully exploit CLIP’s zero-shot classification capabilities in video understanding. Fourth, existing approaches inadequately align cross-modal features by global matching, which disregards the temporal coupling between visual and semantic features.
Motivated by these observations, we introduce Task-Adapter++, a framework that optimally caters to few-shot action recognition (FSAR) through parameter-efficient dual adaptation. Our method comprehensively incorporates task-specific information during video feature extraction, and preserves semantic order information throughout both semantic modeling and cross-modal alignment.

Specifically, to address the first drawback of full fine-tuning, we turn to the recently emerged parameter-efficient finetuning (PEFT) approach. Originating from the field of NLP, PEFT~\cite{houlsby2019parameter,he2021towards,lester2021power} aims to adapt large pre-trained models to downstream applications without fine-tuning all the parameters of the model, while attaining performance comparable to that of full fine-tuning. Inspired by these works, we argue such methods, which only fine-tune a small number of additional adapter parameters, are particularly well-suited for few-shot action recognition, since they balance the preservation of generalizable pre-trained knowledge with the integration of domain-specific cues, enabling more robust and data-efficient video understanding. 

As for the second drawback, existing adapter methods in video domain mainly focus on temporal modeling and lack the consideration of enhancing task-specific features for the few-shot learning tasks. Regarding human cognition, a common consensus is that the most discriminative features vary for different target tasks. For example, in distinguishing between “High Jump” and “Long Jump”, the body movements performed by the actor is more crucial, whereas in distinguishing between “High Jump” and “Pole Vault”, it is more significant whether the actor is holding a long pole. Extracting task-specific features has been a key issue in metric based few-shot image classification ~\cite{ye2020few,li2019finding}, while neglected in FSAR. Although a few works~\cite{wang2022hybrid, liu2022task} consider task-specific adaptation, they solely apply task-specific methods at the level of well-extracted features, which hinders the discovery of the most discriminative information for a given few-shot learning task. In this work, we propose a novel adapter-based method that brings the capability of task-specific adaptation directly into the process of video feature extraction. As shown in Fig.~\ref{fig:fine-tune} (c), by reusing the frozen self-attention blocks with learnable adapters to perform cross-video task-specific attention, our approach dynamically highlights the most discriminative task-relevant patterns within each given task.

Thirdly, motivated by the observation that real-world events inherently follow stage-specific sub-actions, we leverage LLMs to generate and 
structurally link temporally ordered sub-actions that describe an action’s beginning, process and end. Unlike prior text-based enhancements that regard descriptions as static constructs, our approach enforces sequential dependencies between sub-actions, aligning semantic sub-action with the temporal structure of videos. Furthermore, we propose semantic order adaptation, a novel method that explicitly models the sub-action processes underlying complex actions through language, as displayed in Fig.~\ref{fig:fine-tune} (d). This enables the semantic branch to provide fine-grained, order-aware context for the subsequent matching.

Finally, when matching visual and semantic features, to overcome the limitations of global matching and fully take advantage of the temporal coupling between visual and semantic features, we propose a novel fine-grained cross-modal alignment strategy. This strategy aligns video features with semantic features across different stages. Specifically, after fine-grained temporal modeling, which is realized by an attention mechanism, we segment the visual features to directly correspond with the semantic features representing the beginning, process, and end stages. As shown in Fig.~\ref{fig:fine-tune} (d), the final cross-modal matching score is derived from this fine-grained, segment-wise alignment across different temporal stages.

\begin{figure*}[h]
\centering
  \includegraphics[width=.9\textwidth]{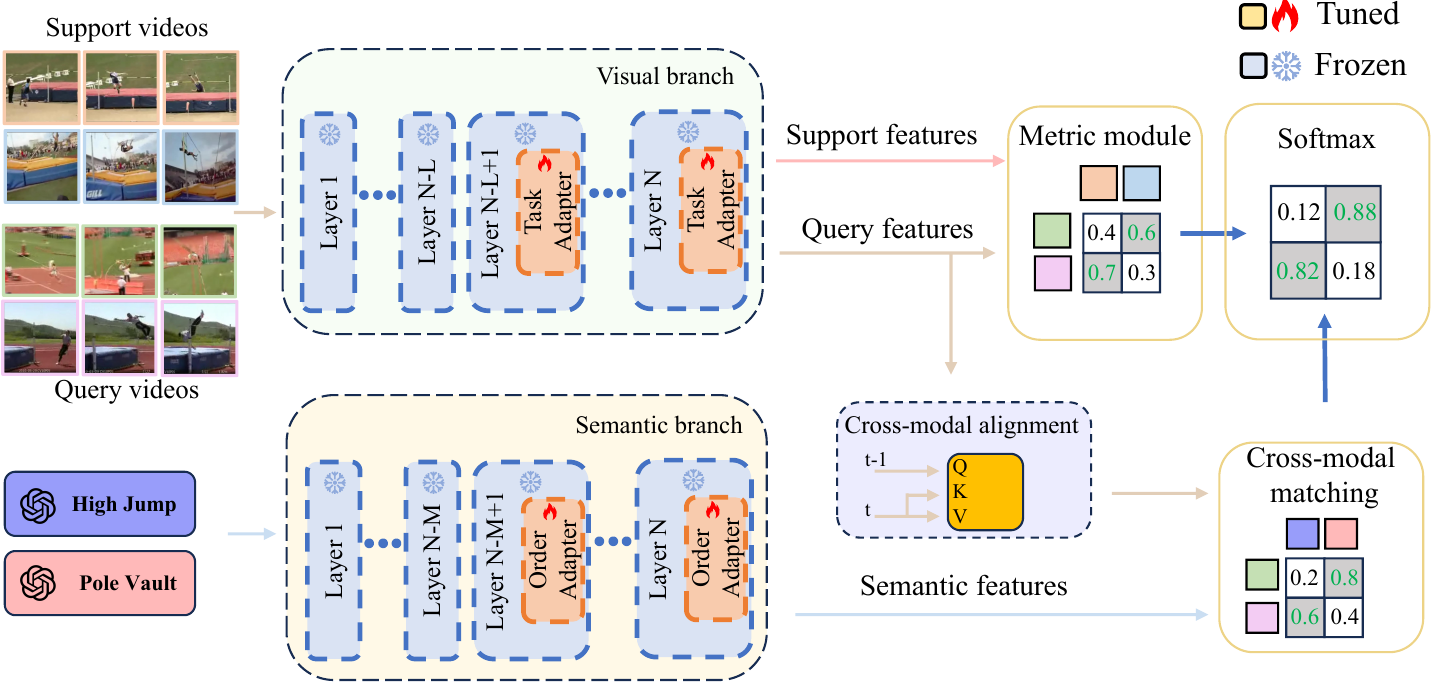}
  \caption{Illustration of Task-Adapter++. We first augment the labels with sub-action descriptions using LLMs, and then independently integrate task adapters and semantic order adapters into the visual and semantic branches. After feature extraction, the support and query features are input into a metric module to compute classification scores, while the query features are further processed through a cross-modal alignment module prior to cross-modal video-text matching. The final score is obtained by combining the outputs from video matching and video-text matching.
  }
\label{fig:framework}
\end{figure*}

As a whole, our framework consists of two branches and computes the matching score accordingly: a visual branch with an image encoder enhanced with task-specific adaptations, and a semantic branch with a text encoder incorporating semantic order adaptations. As shown in Fig.~\ref{fig:framework}, the visual matching score is computed based on the query video features and the support video prototypes extracted by the visual branch, while the cross-modal matching score is derived from the semantic features extracted by the semantic branch and the query video features after cross-modal alignment. The final score is obtained by multiplying these two parts.

Our contributions can be summarized as follows: 
\begin{itemize}
\item 
We introduce a novel parameter-efficient dual adaptation method called Task-Adapter++, which harnesses the unique advantages from both visual and semantic aspects with adapters, effectively alleviating the issues of catastrophic forgetting and overfitting induced by full fine-tuning.

\item 
To the best of our knowledge, we are the first to propose performing task-specific adaptation during the process of spatio-temporal feature extraction, which significantly enhances the discriminative features specific to the given few-shot learning task.

\item 
We extend the semantic information with dynamic sub-action descriptions using LLMs and further model the sequential relationship of actions by semantic order adaptation. After that, we propose a fine-grained cross-modal alignment method which helps matching visual features with semantic features by stages.

\item 
Extensive experiments fully demonstrate the superiority of our Task-Adapter++, establishing new state-of-the-art results on five standard few-shot action recognition benchmarks, and validating the effectiveness of each proposed component.

\end{itemize}

In this paper, we extend our previously published conference paper in MM'24~\cite{cao2024ta} in the following aspects. (i) We propose a novel dual adaptation method for few-shot action recognition, which fine-tunes the adapters concurrently in both image and text encoders. (ii) Benefiting from the LLMs, we extend the labels to more comprehensive and fine-grained sub-action descriptions. This significantly enriches the semantic information at the temporal level. (iii) We further propose a novel adaptation strategy for the semantic branch, namely semantic order adapter, which better models the sequential relationships of sub-actions. (iv) We introduce a fine-grained cross-modal alignment strategy ensuring visual features reside in the same temporal stage as semantic features for better cross-modal feature matching. (v) We conduct comprehensive ablation studies to demonstrate the effectiveness of each proposed component of our Task-Adapter++. The underlying mechanism of each module is further elucidated, providing deeper insights into their contributions. (vi) We significantly boost the performance over the previous version, with an improvement up to 3.4\% on 5 benchmarks.

\section{Related Work}\label{section2}
\subsection{Few-shot Learning}
Few-shot learning aims to learn a model that can quickly adapt to unseen classes using only a few labeled data. Existing works can be divided into three categories: augmentation-based~\cite{mehrotra2017generative, wang2018low}, metric-based~\cite{ye2020few,cao2022learning,snell2017prototypical, vinyals2016matching, xie2022joint, DONG2025111122,LI2024110736,ZHOU2024110790} and optimization-based methods~\cite{finn2017model, xu2020metafun}. 
Augmentation-based methods use auxiliary data to enhance the feature representations or generate extra samples~\cite{mehrotra2017generative} to increase the diversity of the support set. Metric-based methods focus on learning a generalizable feature extractor and classify the query instance by measuring the feature distance. 
Optimization-based methods aim to learn a good network initialization which can fast adapt to unseen classes with minimal optimization steps. 
Among these approaches, metric-based methods are commonly adopted thanks to its simplicity and superior performance. Additionally, task-specific learning has been explored in image-domain metric-based works~\cite{li2019finding,requeima2019fast,bateni2020improved,ma2024cross} to achieve better few-shot learning performance. FEAT~\cite{ye2020few} introduces set-to-set function to obtain task-specific visual features for the few-shot learning tasks. Ta-Adapter~\cite{ZHANG2024TA} introduces a novel approach that enhances few-shot image classification by incorporating task-aware CLIP encoders and a collaborative prompt learning strategy, demonstrating superior performance. However, these methods all introduce task-specific modules at the level of well-extracted image features, which is insufficient for more complex video features due to the additional temporal dimension.
In this work, we follow the metric-based paradigm and perform a novel adaptation during the feature extraction process to achieve more comprehensive adaptation to video data. 

\subsection{Few-shot Action Recognition}
In few-shot action recognition, most of the previous works~\cite{perrett2021temporal,li2022ta2n,zhang2020few,zhu2021few,lu2021few,zhu2021closer,thatipelli2022spatio,xing2023revisiting} fall into the metric-based methods. To solve the problem of temporal misalignment when comparing different videos, a wide series of temporal alignment modules are proposed. OTAM~\cite{cao2020few} improves the dynamic time warping algorithm to preserve the temporal ordering inside the video feature similarities. TRX~\cite{perrett2021temporal} proposes to take the ordered tuples of frames as the least matching unit instead of single frames. CMOT~\cite{lu2021few} uses the optimal transport algorithm to find the best matching between two video features. HyRSM~\cite{wang2022hybrid} treats the sequence matching problem as a set matching problem for better alignment. Except for the aforementioned works, there are also studies focusing on spatial relations~\cite{zhang2023importance} and multi-modal fusion~\cite{fu2020depth,wang2021semantic,wanyan2023active}. However, there is still a lack of research on task-specific learning in few-shot action recognition. In this work, we emphasize the vital importance of task-specific adaptation in identifying the most discriminative spatio-temporal features within the provided video samples.

\subsection{Parameter-Efficient Fine-Tuning}
Parameter-Efficient Fine-Tuning (PEFT) is derived from NLP~\cite{houlsby2019parameter,he2021towards,lester2021power}, aiming to fine-tune large pre-trained models on downstream tasks in an efficient way while achieving comparable performance to full fine-tuning. CLIP-Adapter~\cite{gao2023clip} adds a learnable bottleneck layer behind the frozen CLIP backbone to learn new features for downstream tasks. AIM~\cite{yang2023aim} reuses the frozen pre-trained attention layer and applies it to the temporal dimension of the input video. SAFE~\cite{10646624} applies PEFT to semantic awareness that enhance the comprehension ability of image classification. However, in the FSAR task, no work has ever considered extracting task-specific visual features and modeling the sequential information of sub-actions using PEFT. In this work, we propose a novel PEFT method to acquire the most discriminative information and order information on two branches of CLIP for FSAR. 

\section{Method}\label{section3}
\subsection{Problem definition}
Few-shot action recognition aims to learn knowledge from base classes and classify human action on new classes with limited data. To achieve this goal, we adopt an episode-based training manner, which can be performed as $C$-way $K$-shot. The video datasets are partitioned into nonoverlapping training and testing splits, denoted as $\mathcal{D}_{tr}$ and $\mathcal{D}_{ts}$, respectively. In each episode, we randomly sample a $C$-way $K$-shot task, which has $C$ action classes and $K$ labeled samples in each class, from $\mathcal{D}_{tr}$ for each episode during training. All the $C\times K$ video samples form the support set $\mathcal{S}_{tr}=\{x_i\}_{i=1}^{C\times K}$. Additionally, there is a query set $\mathcal{Q}_{tr}=\{x_i\}_{i=1}^{Q_{tr}}$ from the same $C$ action classes. The learning objective of each task is to classify each query video $q\in \mathcal{Q}_{tr}$ into one of the $C$ classes. In test stage, we sample $C$-way $K$-shot task from $\mathcal{D}_{ts}$ and classify the query video $q \in \mathcal{Q}_{ts}$ into one of the $C$ classes with frozen parameters. 

\begin{figure*}[h]
\centering
  \includegraphics[width=.9\textwidth]{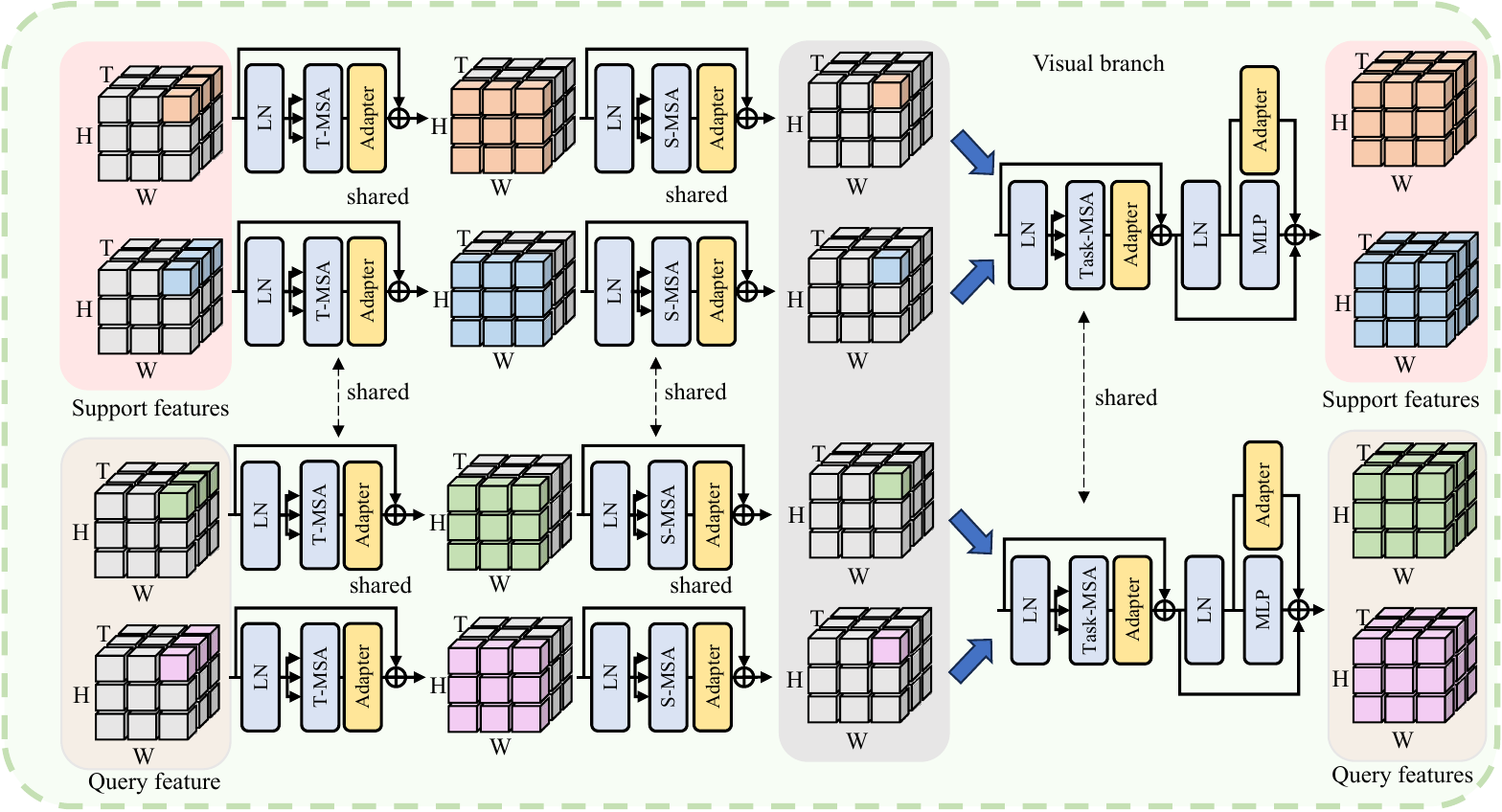}
  \caption{Illustration of Task-Adapter in visual branch. For task-specific adaptation, we introduce task adapter after T-MSA and S-MSA to enhance the task-specific information for the few-shot action recognition. This part illustrates the computational process of the image encoder given a 2-way 1-shot learning task with two query videos in the query set.
  }
\label{fig:task-adapter}
\end{figure*}

\subsection{Framework}
The architecture of the whole Task-Adapter++ is shown in Fig.~\ref{fig:framework}, which demonstrate our dual adaptation method. 
The visual branch is shown in Fig.~\ref{fig:task-adapter} where we introduce task-specific adaptation to pay attention to the most discriminative video information in a given task. And Fig.~\ref{fig:order-adapter} shows the semantic branch which model the sequential information between sub-actions by our proposed semantic order adaptation. The classification score consists of two parts: a metric score derived from the support and query features, and a cross-modal matching score computed from the query and semantic features. The details of each component are described as follows.

\begin{figure*}[h]
  \setlength{\abovecaptionskip}{-5pt}
  \setlength{\belowcaptionskip}{-5pt}
  \centering
  \includegraphics[width=0.35\textwidth]{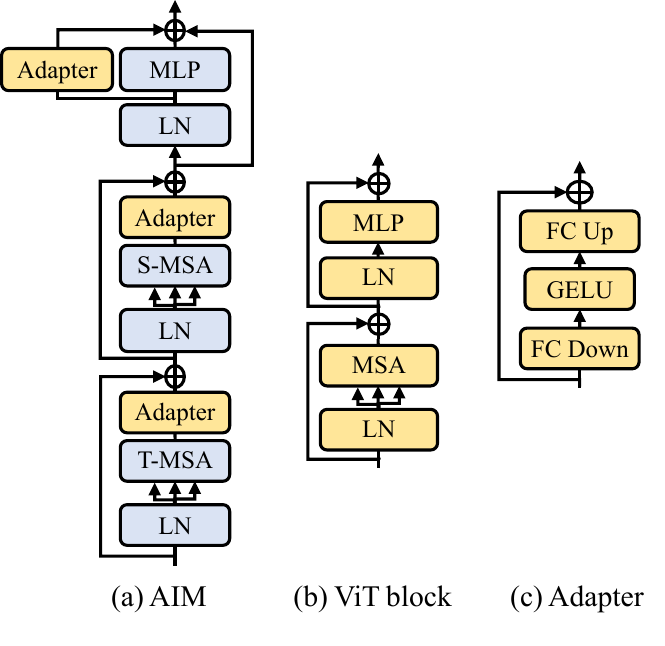}
  \caption{
  AIM (a) adapts the standard ViT bock (b) by freezing the original pre-trained model (outlined with a blue background) and adding tunable Adapters (c) individually for temporal adaptation, spatial adaptation and joint adaptation.  }

  \label{fig:adapter}
\end{figure*}

\subsection{Adapting ViT to Action Recognition}\label{3.3}
Given an RGB video sample $v\in \mathbb{R}^{T\times H\times W\times 3}$, a vanilla ViT takes the temporal dimension $T$ as the batch dimension $B$, i.e., there is no exchange of information between frames. Specifically, ViT divides each frame into $N=HW/P^2$ patches, where $P$ is the patch size, and transforms the patches into $D$ dimensional patch embeddings $v_p\in \mathbb{R}^{T\times N\times D}$. A learnable [class] token is prepended to the patch embeddings $v_p$ as $v_0\in \mathbb{R}^{T\times (N+1)\times D}$. To retain position information, a learnable positional embedding $E_{pos}\in \mathbb{R}^{(N+1)\times D}$ is added to $v_0$ as $z_0=v_0+E_{pos}$, where the $E_{pos}$ is broadcast to $T$ dimension to be shared among different frames and $z_0$ is the final input of the stacked ViT blocks. Each ViT block consists of LayerNorm (LN), multi-head self-attention (MSA) and MLP layers, with residual connections after each block, as shown in Fig.~\ref{fig:adapter} (b). The computation process during each ViT block can be given by:
\begin{align}
        z_l^\prime &= {\rm MSA}({\rm LN}(z_{l-1}))+z_{l-1}\label{eq:1}\\
    z_l &= {\rm MLP}({\rm LN}(z_l^\prime)) + z_l^\prime 
\label{eq:2}
\end{align}
where $z_{l}$ denotes the output of the $l$-th ViT block and all the MSAs are computed on patch dimension to learn the relationship of different spatial parts of the original frame. The [class] tokens of each frame from the last ViT block are concatenated as the final video feature $F_v\in \mathbb{R}^{T\times D}$ which is fed into the classifier and used to calculate the final classification score.

Inspired by PEFT techniques, AIM tries to adapt pre-trained image model to video action recognition with minimal fine-tuning by freezing the original model and only fine-tuning additionally introduced adapters shown in Fig.~\ref{fig:adapter} (c). Moreover, AIM proposes to reuse the pre-trained MSA layer to do temporal modeling which is deficient in the vanilla ViT. As shown in Fig.~\ref{fig:adapter} (a), a shared frozen temporal MSA (T-MSA) layer is prepended to the original spatial MSA (S-MSA) sequentially to exchange the temporal information between frames for temporal modeling, which can be achieved easily by permuting the patch dimension and the temporal dimension. The tunable adapter is added after each MSA layer and parallel to the MLP layer to achieve temporal, spatial and joint adaptation as follows:
\begin{align}
    z_l^t &= {\rm Adapter}({\rm T \mbox{-} MSA}({\rm LN}(z_{l-1})))+z_{l-1} \\
    z_l^s &= {\rm Adapter}({\rm S\mbox{-}MSA}({\rm LN}(z_l^t)))+z_l^t \\
    z_l &= {\rm Adapter}(({\rm LN}(z_l^s)) + {\rm MLP}({\rm LN}(z_l^s)) + z_l^s
  \label{eq:aim}
\end{align}
\noindent where $z_l^t$, $z_l^s$ and $z_l$ individually stands for temporal adapted, spatial adapted and jointly adapted output from the $l$-th ViT block. However, directly transferring this method to few-shot action recognition is inadequate due to the lack of considering the relationship between different videos within the few-shot learning task and the information gap between seen classes and unseen classes.

\subsection{Adapting ViT to Task-specific FSAR}\label{3.4} FSAR studies how to classify unseen class query videos with only a limited number of labeled samples available. Although the temporal adaptation proposed by AIM gives vanilla ViT the ability of temporal modeling, it remains unaddressed that the ViT model could not attend to different videos and enhance the task-specific discriminative features. 

Therefore, we describe how we empower the frozen ViT with the ability to adapt embeddings with task-specific information for few-shot action recognition. As described in Section~\ref{3.3}, we first map all the videos in the whole sampled task (both support set and query set) into their patch embeddings by Eq.~\eqref{eq:1} and Eq.~\eqref{eq:2}, then we get the input features of the first ViT block $\mathcal S_0 = [x_0^1, ..., x_0^{CK}]$ and $\mathcal Q_0 = [x_0^1, ..., x_0^{Q}] $, where $x_0^i \in \mathbb{R}^{T\times(N+1)\times D}$ denotes the patch embeddings of the $i$-th video both for the support set and query set. Then, these patch embeddings are fed into several frozen ViT blocks to exchange spatial information between different tokens. It should be noted that we only introduce tunable adapters in the last \textit{L} ViT blocks, which is a choice based on empirical study. We will further analyze the impact of this decision in the ablation experiment. Starting from the (\textit{N-L+1})-th layer, we introduce tunable adapter layers into the original model. Specifically, besides T-MSA and S-MSA, we further reuse the frozen MSA layer as Task-MSA after the temporal and spatial adaptation to perform task-specific self-attention across the tokens at the same spatio-temporal location in different videos (see Fig.~\ref{fig:task-adapter}). Take the support features as an example:
\begin{align}
    \mathcal S_l^t &= {\rm Adapter}({\rm T \mbox{-} MSA}({\rm LN}(\mathcal S_{l-1})))+\mathcal S_{l-1}\\
    \mathcal S_l^s &= {\rm Adapter}({\rm S\mbox{-}MSA}({\rm LN}(\mathcal S_l^t)))+\mathcal S_l^t \\
    \mathcal S_l^{task} &= {\rm Adapter}({\rm Task\mbox{-}MSA}({\rm LN}(\mathcal S_l^s)))+\mathcal S_l^s \\
    \mathcal S_l &= {\rm Adapter}({\rm LN}(\mathcal S_l^{task})) + {\rm MLP}({\rm LN}(\mathcal S_l^{task})) + \mathcal S_l^{task}
  \label{eq:task}
\end{align}
\noindent where $\mathcal{S}_{l-1},\mathcal{S}_l \in \mathbb{R}^{CK\times T \times(N+1)\times D}$ individually denotes the input and output support features of the $l$-th layer ($l \in \{N-L+1, N-L+2, ..., N\}$).

The query features can be computed in the same way. To avoid exposing the target support class features to the specific query video feature, we isolate the information interaction between the support set and query set.

Finally, we concatenate the [class] tokens of each frame from the last ViT block to get the support set video representations $\mathcal{F}_{\mathcal{S}} \in \mathbb{R}^{CK \times T \times D}$ and the query set video representations $\mathcal{F}_{\mathcal{Q}}\in \mathbb{R}^{Q \times T \times D}$. The video metric score is calculated based on the support prototypes and query features.

For the support video features $\mathcal{F}_{\mathcal{S}}$, the support video prototype can be denoted as: 
\begin{align}
  \hat{\mathcal{F}}_{\mathcal{S}}^c = \frac{1}{K}\sum^K_i{\mathcal{F}_{\mathcal{S}}[c][i]}
  \label{eq:prototype}
\end{align}
\noindent where $\hat{\mathcal{F}}_{\mathcal{S}}^c \in \mathbb{R}^{T\times D}$ denotes the class prototype of the $c$-th support class. We directly average all the support video features as the class prototype for each class. 

To calculate the classification probability, we measure the distance between the query features and the class prototypes by an arbitrary metric module, e.g., TRX~\cite{perrett2021temporal}, OTAM~\cite{cao2020few} and Bi-MHM~\cite{wang2022hybrid}, which can formulated by:
\begin{align}
  \mathcal{P}^c_i = {\rm Metric}(\mathcal{F}_{\mathcal{Q}}[i], \hat{\mathcal{F}}_{\mathcal{S}})
  \label{eq:inference}
\end{align}
\noindent where $\mathcal{P}^c_i$ is the probability of the $c$-th class for the $i$-th query video in query set and $\hat{\mathcal{F}}_{\mathcal{S}} = [\hat{\mathcal{F}}_{\mathcal{S}}^1, \hat{\mathcal{F}}_{\mathcal{S}}^2, ..., \hat{\mathcal{F}}_{\mathcal{S}}^c] \in \mathbb{R}^{C\times T\times D}$ stands for all the class prototypes of the whole support set. 

In our experiments of different metric modules, we will demonstrate that our proposed Task-Adapter++ consistently improves the performance of any metric methods. Besides that, owning to the superiority of our task-specific adaptation during feature extraction, excellent performance can be achieved simply by averaging the frame-to-frame cosine similarity without complex metric measurements.

\begin{figure*}[h]
\centering
  \includegraphics[width=.9\textwidth]{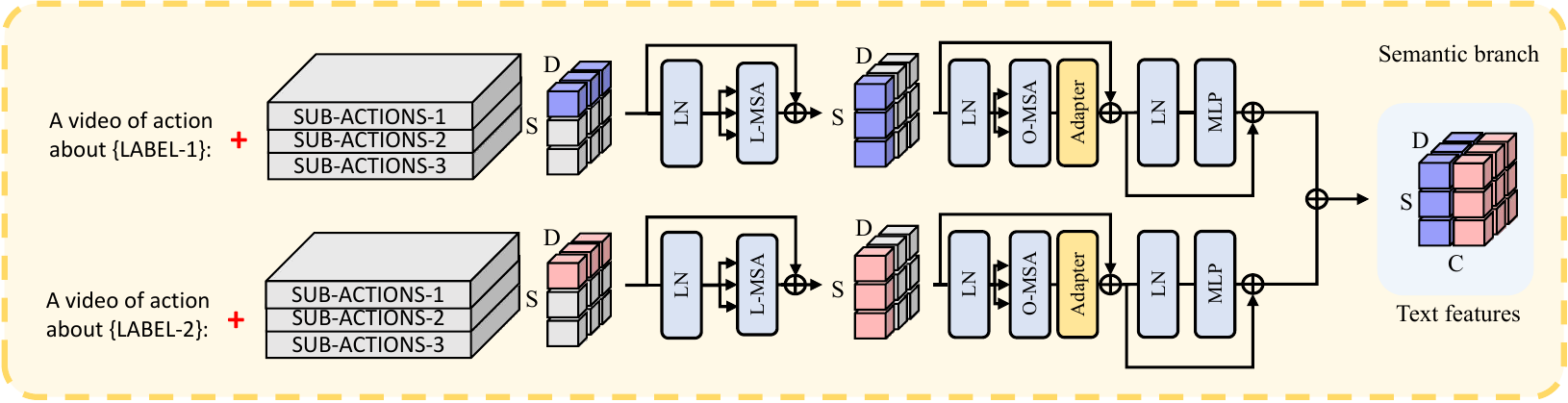}
  \caption{Illustration of order adaptation in semantic branch. For semantic order adaptation, we introduce semantic order adapter after MSA to model the sequential information of sub-actions. This part illustrates the computational process of the text encoder given a 2-way 1-shot learning task.
  }
\label{fig:order-adapter}
\end{figure*}
\subsection{Semantic Order Adaptation} 
Considering that existing works overlook the fine-grained sequential modeling in the semantic feature extraction process, we propose the semantic order adaptation. Essentially, our primary task is to construct a semantic corpus that contains dynamic descriptions of different actions. We prompt the LLMs to generate fine-grained sub-actions with following instruction: ``\textit{Given an action label} \{LABEL\}, \textit{describe three stages of the action}.'' The LLMs generate corresponding descriptions of the beginning, process, and end for each input action label. For example, if the input label is ``Long Jump'', the corresponding output is a list delineates the beginning, process, and end of the action as follows: [``Run up for momentum.'', ``Take off with a powerful leap.'', ``Land and maintain balance.'']

After obtaining the fine-grained descriptions as above, we concatenate the original label with the three sub-actions respectively to form three prompts, which are then used as input for the semantic branch of CLIP, such as ``\textit{A video of action about} \{LABEL\}: \{SUB-ACTIONS-1\}'', ``\textit{A video of action about} \{LABEL\}: \{SUB-ACTIONS-2\}'' and ``\textit{A video of action about} \{LABEL\}: \{SUB-ACTIONS-3\}''.

As shown in Fig.~\ref{fig:order-adapter}, to enhance the representation of sequential relationships between sub-actions, inspired by PEFT, we freeze the text encoder and introduce a tunable adapter after the last $M$-layer to achieve semantic order adaptation. Specifically, we reuse the frozen MSA layers as O-MSA, which could model the sequential information across the three sub-actions as follows:
\begin{align}
    \mathcal{T}_{j'}^c &= {\mathrm{MSA}}({\mathrm{LN}}(\mathcal{T}_{j-1}^c))+\mathcal{T}_{j-1}^c\\
    \mathcal {T}_{j''}^c &= {\rm Adapter}({\rm O\mbox{-}MSA}({\rm LN}(\mathcal {T}_{j'}^c))) + \mathcal{T}_{j'}^c\\
    \mathcal {T}_{j}^c &= {\rm MLP}({\rm LN}(\mathcal {T}_{j''}^c)) + \mathcal {T}_{j''}^c
  \label{eq:order}
\end{align}
\noindent where the output semantic feature of class $c$ in the $j$-th ViT block  ($j \in \{N-M+1, N-M+2, ..., N\}$) is denoted as $\mathcal{T}_j^c = \{\mathcal{T}_j^{c_1}, \mathcal{T}_j^{c_2}, \mathcal{T}_j^{c_3}\} \in \mathbb{R}^{l \times 3\times D} $, and $l$ presents the sequence length in text encoder. 

Finally, for each action within a task, we select the end-of-text token from the text features in the last block and concatenate them to form the output semantic feature for cross-modal matching. 
\begin{figure}[h]
\centering
  \includegraphics[width=.45\textwidth]{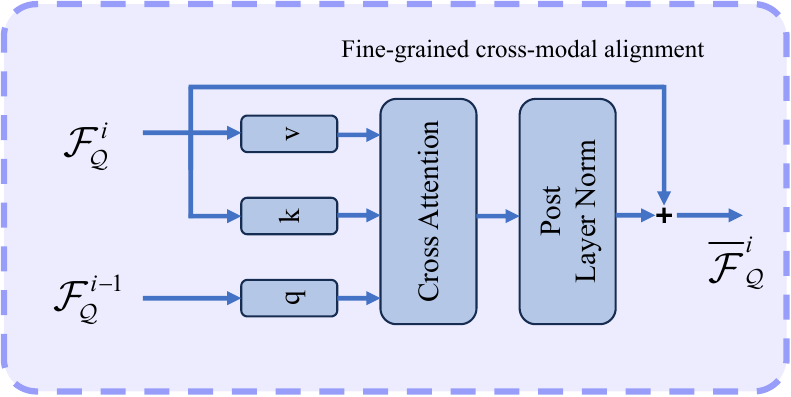}
  \caption{The structure of our fine-grained cross-modal alignment. We use cross-attention for fine-grained temporal modeling, where the ($i\mbox{-}1$)-th frame serves as $q$ and the $i$-th frame serves as $k$ and $v$.
  }
\label{fig:align}
\end{figure}
\subsection{Fine-grained Cross-modal Alignment}\label{3.6}
Considering that videos contain richer spatio-temporal information while semantics are more abstract, mapping videos into the semantic space aligns more naturally with the principles of information extraction. Thus, we propose a fine-grained cross-modal alignment framework that models the temporal relationships between adjacent frames. In this framework, visual features are aligned to correspond with the same temporal stages as their semantic descriptions. As illustrated in Fig.~\ref{fig:align}, our method enables more precise alignment by modeling frame-to-frame dynamics in the temporal context with cross-attention:
\begin{align}
    \Bar{\mathcal{F}}_{Q}^{i} &= \text{Cross-Attention}(\mathcal{F}_{Q}^{i-1},\mathcal{F}_{Q}^{i})+\mathcal{F}_{Q}^{i-1}
  \label{eq:smoother}
\end{align}
\noindent where $\mathcal{F}_{Q}^{i} \in \mathbb{R}^{Q\times D}$ denotes the $i$-th frame of the query's visual feature set, and for ease of understanding we define the aligned features as $\Bar{\mathcal{F}}_{Q}\in \mathbb{R}^{Q\times (T-1)\times D}$. By default, $T$ is set to 8. As no padding is applied along the temporal dimension, the length of the aligned features is reduced by one compared to the original input sequence..

For the $q$-th aligned visual features $\Bar{\mathcal{F}}_{\mathcal{Q}_q}$ in the query set, we divide it into three temporal segments, corresponding to the three different sub-actions, referred to as:
\begin{equation}
    \Bar{\mathcal{F}}_{\mathcal{Q}_q}=[\frac{1}{3}\sum^3_{i=1} \Bar{\mathcal{F}}_{\mathcal{Q}_q}^i,\frac{1}{3}\sum^5_{i=3} \Bar{\mathcal{F}}_{\mathcal{Q}_q}^i,\frac{1}{3}\sum^7_{i=5} \Bar{\mathcal{F}}_{\mathcal{Q}_q}^i]\in \mathbb{R}^{3\times D}
\end{equation}

\begin{figure*}[h]
\centering
  \includegraphics[width=.9\textwidth]{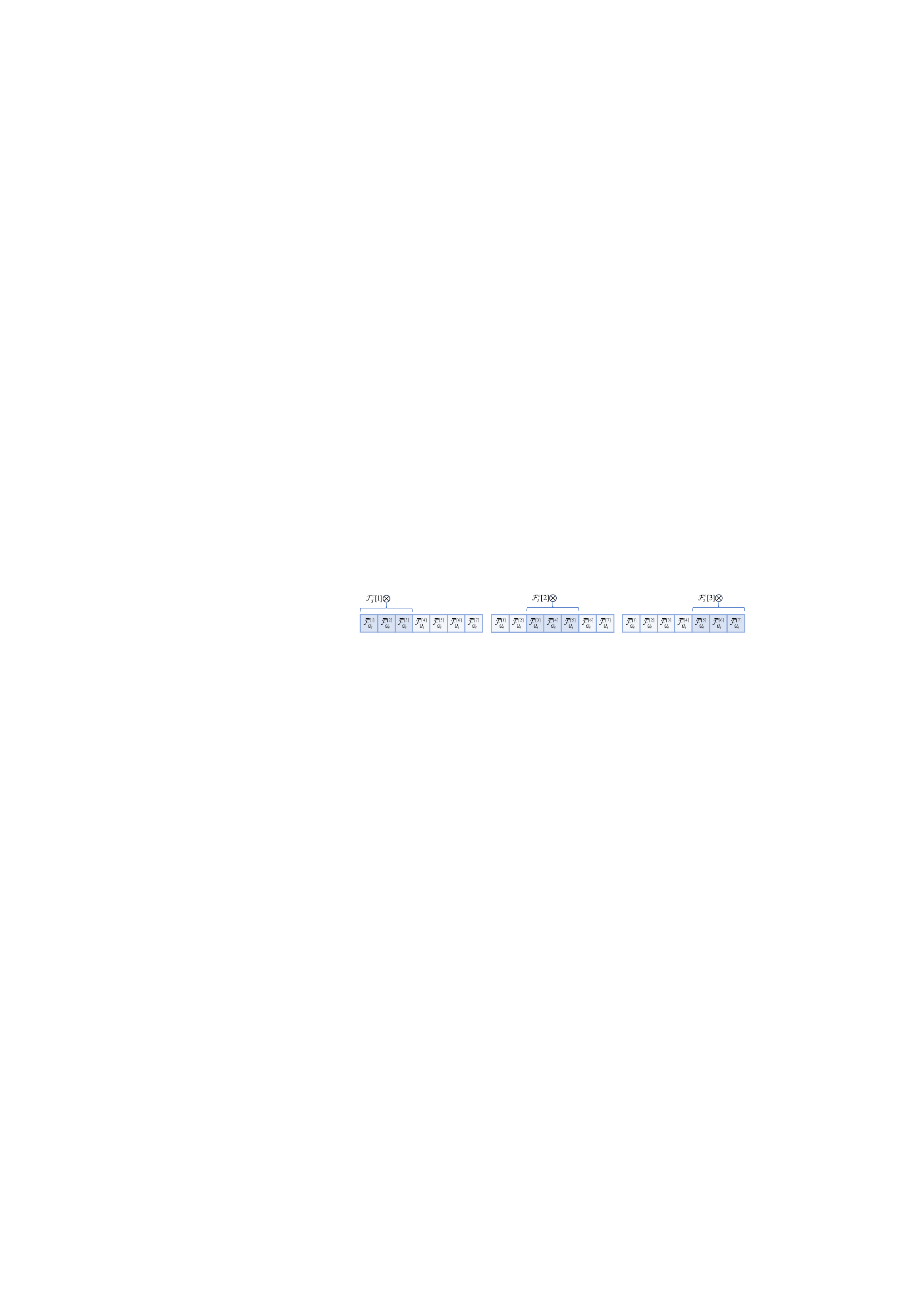}
  \caption{The stage-wise illustration of the cross-modal feature matching process. The aligned query features are divided into three segments and matched with the features of three sub-actions respectively.
  }
\label{fig:3stage}
\end{figure*}

Then, the probability distribution is calculated based on the cosine similarity between the semantic feature $\mathcal{F}_\mathcal{T}$ and the aligned query feature segment $\Bar{\mathcal{F}}_{\mathcal{Q}_q}$, which can be formulated as:

\begin{equation}
\dot{\mathcal{P}}_q^c
=\frac{1}{3}\sum_{s=1}^3 \text{Cos}(\Bar{\mathcal{F}}_{\mathcal{Q}_q}[s],\mathcal{F}_\mathcal{T}^c[s])
\end{equation}

\noindent where $\dot{\mathcal{P}}_q^c$ is the probability of the $c$-th class for the $q$-th video in query set. $\Bar{\mathcal{F}}_{\mathcal{Q}_q}[s]$ means the $s$-th query feature segment for the $q$-th query video and $\mathcal{F}_\mathcal{T}^c[s]$ means the $s$-th sub-action features of class $c$. The process is also illustrated in Fig.~\ref{fig:3stage}. 

Notably, instead of averaging temporal video features for global matching, we divide them into three temporal segments to facilitate fine-grained cross-modal alignment. This design enables the visual and textual features to be matched at each corresponding stage individually, resulting in more precise and discriminative matching result.

\subsection{Calculation of Classification Results}
The final classification score consists of two components: the video matching score $\mathcal{P}$, as demonstrated in Section~\ref{3.4}, which measures the similarity between support and query features based on video prototypes; and the video-text matching score $\mathcal{\dot{P}}$, introduced in Section~\ref{3.6}, which evaluates the alignment between semantic features and query features. 
These two scores are combined via element-wise multiplication to jointly consider both visual and semantic similarities. Thus, the final score is calculated in the form of a product:
\begin{equation}
    \text{Score} = \mathcal{P} \times \dot{\mathcal{P}}
\end{equation}

The similarity scores for each task are treated as logits and used to compute the cross-entropy loss, enabling gradient backpropagation during training. Notably, we only fine-tune the parameters of the additional adapter layers, while keeping the backbone of the pre-trained model frozen. During testing, the entire model, including the adapters, is fixed, and the highest-scoring class for each query video is selected as the final prediction. 

\begin{table}[h]
  \begin{center}
    \caption{Comparison with current SOTA few-shot action recognition methods on temporal-related datasets (SSv2-Small and SSv2-Full) with 5-way 1-shot, 5-way 3-shot and 5-way 5-shot settings. The best results are highlighted in bold.}
  \label{tab:sota1}
  \resizebox{0.85\textwidth}{!}{

        \begin{tabular}{l c c c c c c c c c c c}
          \toprule
      \multirow{2}{4em}{Method}&\multirow{2}{4em}{Reference}&\multirow{2}{3.5em}{Backbone}&\multirow{2}{3.5em}{Pretrain}&\multirow{2}{5em}{Fine-tuning}&\multicolumn{3}{c}{SSv2-Small}&\multicolumn{3}{c}{SSv2-Full}\\ 
      & & & & &1-shot & 3-shot & 5-shot & 1-shot & 3-shot & 5-shot\\
      \midrule
      CMN~\cite{zhu2018compound} & ECCV'18 & ResNet50 & IN-21K & Full Fine-tuning & 34.4 & 42.5 & 43.8 & 36.2 & - & 48.9\\ 

      OTAM~\cite{cao2020few}& CVPR'20 & ResNet50 & IN-21K & Full Fine-tuning &- & - & - & 42.8 & 51.5 & 52.3  \\  
    
      TRX~\cite{perrett2021temporal}& CVPR'21 &ResNet50 & IN-21K & Full Fine-tuning &36.0 & 51.9 & 59.1 & 42.0 & 57.6 & 64.6  \\  


      STRM~\cite{thatipelli2021spatio} & CVPR'22&ResNet50 & IN-21K & Full Fine-tuning & 37.1 & 49.2 & 55.3 & 43.1 & 59.1 & 68.1 \\  
      

      HyRSM~\cite{wang2022hybrid}& CVPR'22&ResNet50 & IN-21K & Full Fine-tuning &40.6 & 52.3 & 56.1 & 54.3 & 65.1 & 69.0\\ 
      HCL~\cite{hcl} & ECCV'22 & ResNet50 & IN-21K & Full Fine-tuning & 38.7 & 49.1 & 55.4 & 47.3 & - & - \\
      MoLo~\cite{wang2023molo}& CVPR'23&ResNet50 & IN-21K & Full Fine-tuning &41.9 & - & 56.2 & 55.0 & - & 69.6 \\ 
      HyRSM++~\cite{WANG2024110110} & PR'24 &ResNet50 & IN-21K & Full Fine-tuning &42.8 & 52.4 & 58.0 & 55.0 & 66.0 & 69.8 \\ 
      MASTAF~\cite{liu2023mastaf} & WACV'23 & ViViT & JFT & Full Fine-tuning & 45.6 & - & - & 60.7 & - & -\\
      
      SA-CT~\cite{zhang2023importance} & MM'23 & ViT-B & IN-21K & Full Fine-tuning & - & - & - & - & - & 66.3 \\
      CLIP-FSAR~\cite{wang2023} & IJCV'23& ViT-B & CLIP & Full Fine-tuning &54.6 & - & 61.8 & 62.1 & - & 72.1\\ 
      
      MA-FSAR~\cite{xing2023multimodal} & ArXiv'23 &ViT-B & CLIP & PEFT &59.1 & - & 64.5 & 63.3 & - & 72.3\\ 
      
      TSAM~\cite{Li_Liu_Wang_Yu_2025} & AAAI'25 & ViT-B & CLIP & PEFT & 60.5 & - & 66.7 & 65.8 & - & 74.6 \\

      \hline
      \textbf{Task-Adapter} & MM'24 &ViT-B & CLIP & PEFT & 60.2  & 66.1 & 70.2 & 71.3 & 71.5 & 74.2\\
      \textbf{Task-Adapter++} &- &ViT-B &CLIP & PEFT & \textbf{63.6} & \textbf{69.1} & \textbf{73.2} & \textbf{72.8} & \textbf{73.0} & \textbf{75.5}\\
      \hline
      \end{tabular}   
}
  \end{center}
\end{table}
\section{Experiments}\label{section4}
\subsection{Experimental Settings}
\textbf{Datasets.} We evaluate our method on five datasets that widely used in few-shot action recognition, namely HMDB51
~\cite{6126543}, UCF101~\cite{soomro2012ucf101}, Kinetics~\cite{carreira2017quo} and SSv2~\cite{goyal2017something}. The first three datasets focus on scene understanding, while the last dataset, SSv2, has been shown to require more challenging temporal modeling~\cite{Tu_2023_ICCV, cao2020few}. Particularly, We use the few-shot split proposed by~\cite{zhang2020few} for HMDB51 and UCF101, with 31/10/10 classes and 70/10/21 classes for train/val/test, respectively. Following~\cite{zhu2018compound}, Kinetics is used by selecting a subset which consists of 64, 12 and 24 training, validation and testing classes. For SSv2, we provide the results of two publicly available few-shot split, i.e., SSv2-Small~\cite{zhu2020label} and SSv2-Full~\cite{cao2020few}. Both SSv2-Small and SSv2-Full contain 100 classes selected from the original dataset, with 64/12/24 classes for train/val/test, while SSv2-Full contains 10x more videos per class in the training set. 

\textbf{Implementation details.} As previously illustrated, on one hand we aim to adapt the image pre-trained ViT model for task-specific few-shot action recognition, on the other hand we use order adaptation to model semantic information of sub-actions. To better demonstrate the model's capabilities, we utilize a CLIP pre-trained ViT model~\cite{radford2021learning} as the backbone. For a fair comparison with the existing works~\cite{wang2023,xing2023multimodal}, we uniformly sample 8 frames for each video, and crop the frames to 224$\times$224 as the input resolution. Before the training, we use GPT-4~\cite{openai2024gpt4technicalreport} to enhance our label information. The semantic corpus is saved as json file without any manual modification. During the training stage, random crop is used for data augmentation. We freeze the original pre-trained model and only fine-tune the introduced Task adapters and order adapters using the SGD optimizer with a learning rate of 0.001. All reported results in our paper are the average accuracy over 10,000 few-shot action recognition tasks.

\begin{table*}[h]
  \begin{center}
    \caption{Comparison with current SOTA few-shot action recognition methods on scene-related datasets (HMDB51, UCF101 and Kinetics) with 5-way 1-shot, 5-way 3-shot and 5-way 5-shot settings. The best results are highlighted in bold.}
  \label{tab:sota2}
  \resizebox{0.97\textwidth}{!}{
        \begin{tabular}{l c c c c c c c c c c c c c c}
          \toprule
      \multirow{2}{4em}{Method}&\multirow{2}{4em}{Reference}&\multirow{2}{3.5em}{Backbone}&\multirow{2}{3.5em}{Pretrain}&\multirow{2}{5em}{Fine-tuning}&\multicolumn{3}{c}{HMDB51}&\multicolumn{3}{c}{UCF101}&\multicolumn{3}{c}{Kinetics}\\ 
      & & & & & 1-shot & 3-shot & 5-shot & 1-shot & 3-shot & 5-shot & 1-shot & 3-shot & 5-shot\\
      \midrule
      CMN~\cite{zhu2018compound} & ECCV'18 & ResNet50 & IN-21K & Full Fine-tuning &- & - & - & - & - & - & 60.5 & 75.6 & 78.9\\ 

      ARN~\cite{zhang2020few} & ECCV'20 & C3D & - & Full Fine-tuning & 44.6 & - & 59.1 & 62.1 & - & 84.8&63.7 & - & 82.4 \\ 
      OTAM~\cite{cao2020few}& CVPR'20 & ResNet50 & IN-21K & Full Fine-tuning & 54.5 & 65.7 & 68.0 & 79.9 & 87.0 & 88.9 & 73.0 & 78.7 & 85.8 \\  
      
      TRX~\cite{perrett2021temporal}& CVPR'21 &ResNet50 & IN-21K & Full Fine-tuning & 53.1 & 67.4 & 75.6 &78.2& 92.4 & 96.1&63.6 &81.8& 85.9 \\  


      STRM~\cite{thatipelli2021spatio} & CVPR'22&ResNet50 & IN-21K & Full Fine-tuning & 52.3 & 67.4 & 77.3 & 80.5 & 92.7 & 96.9 & 62.9 & 81.1  & 86.7 \\  
      

      HyRSM~\cite{wang2022hybrid}& CVPR'22&ResNet50 & IN-21K & Full Fine-tuning &60.3 & 71.7 & 76.0 & 83.9 & 93.0 & 94.7 &73.7 & 83.5 & 86.1 \\ 
       
      HCL~\cite{hcl} & ECCV'22 & ResNet50 & IN-21K & Full Fine-tuning & 59.1 & 71.2 & 76.3 & 82.5 & 91.0 & 93.9 & 73.7 & 82.4 & 85.8\\
      MoLo~\cite{wang2023molo}& CVPR'23&ResNet50 & IN-21K & Full Fine-tuning &60.8 & - & 77.4 & 86.0 & - & 95.5 &74.0 & - & 85.6 \\ 
      MGCSM~\cite{yu2023multi} & MM'23 & ResNet50 & IN-21K & Full Fine-tuning & 61.3 & - & 79.3 & 86.5 & - & 97.1 & 74.2 & - & 88.2\\
      HyRSM++~\cite{WANG2024110110} & PR'24 &ResNet50 & IN-21K & Full Fine-tuning &61.5 & 72.7 & 76.4 & 85.8 & 93.5 & 95.9 &74.0 & 83.9 & 86.4 \\
      MASTAF~\cite{liu2023mastaf} & WACV'23 & ViViT & JFT & Full Fine-tuning & 69.5 & - & - & 91.6 & -& - & - & - & -\\
      
      SA-CT~\cite{zhang2023importance} & MM'23 & ViT-B & IN-21K & Full Fine-tuning & - & - & 81.6 & - & - & 98.0 & - & - & 91.2 \\
      
      CLIP-FSAR~\cite{wang2023} & IJCV'23& ViT-B & CLIP & Full Fine-tuning &77.1 & - & 87.7 & 97.0 & - & 99.1& 94.8 & - & 95.4\\ 
      MA-FSAR~\cite{xing2023multimodal} & ArXiv'23 &ViT-B & CLIP & PEFT &83.4 & - & 87.9 & 97.2 & - &99.2 & 95.7 & - & 96.0\\ 
      TSAM~\cite{Li_Liu_Wang_Yu_2025} & AAAI'25 & ViT-B & CLIP & PEFT & 84.5 & - & 88.9 & 98.3 & - & 99.3 & 96.2 & - & 97.1\\
      \hline
      \textbf{Task-Adapter} & MM'24 &ViT-B & CLIP & PEFT & 83.6 & 88.0 & 88.8 & 98.0 & 98.7 & 99.4& 95.0 & 96.3 & 96.8\\
      \textbf{Task-Adapter++} &- &ViT-B &CLIP & PEFT & \textbf{86.5} & \textbf{90.6} & \textbf{92.0} & \textbf{98.9} & \textbf{99.5} & \textbf{99.7} & \textbf{96.2} & \textbf{97.3} & \textbf{97.5}\\
      \hline
      \end{tabular}   
  }
  \end{center}
\end{table*}
\subsection{Comparison with state-of-the-art methods}
Table~\ref{tab:sota1} and Table~\ref{tab:sota2} presents a comprehensive comparison of our method with recent few-shot action recognition methods on both temporal-related datasets and scene-related datasets. As our Task-Adapter++ is designed for ViT-based architectures, we use CLIP~\cite{radford2021learning} pre-trained ViT to make a fair comparison with recent few-shot action recognition works that use large-scale pretrained models. Of particular note, by comparing the results of works that use CLIP pre-trained models and those that use ImageNet (IN-21K) pre-trained models, we can conclude that the performance of few-shot action recognition indeed benefits from large-scale pretraining, especially for scene-related datasets. 

As shown in Table~\ref{tab:sota1}, our Task-Adapter++ can achieve the best results on SSv2-Small and SSv2-Full, leading previous SOTA method with a large margin, which indicates the superiority of our method on temporal challenging few-shot learning tasks. The results of Task-Adapter compared with MA-FSAR~\cite{xing2023multimodal} on the SSv2 datasets demonstrate the effectiveness of our proposed method for task-specific adaptation, showing a significant improvement of 5.7\% for the 5-shot task on SSv2-Small and 8.0\% for the 1-shot task on SSv2-Full, respectively. Our Task-Adapter have achieved an average improvement of 3.4\% on SSv2-Small and 4.9\% on SSv2-Full compared to the SOTA method MA-FSAR~\cite{xing2023multimodal}.
Our Task-Adapter++ have achieved further improvement on the basis of Task-Adapter, specifically improving 3.1\% on SSv2-Small and 1.4\% on SSv2-Full, benefiting from the semantic order adaptation and fine-grained cross-modal matching ability. 
Compared with full fine-tuning method CLIP-FSAR~\cite{wang2023}, we can see our Task-Adapter++ has a significant overall improvement (from 3.4\% to 11.4\%) with fewer learnable parameters (7.5 M \textit{vs}. 89.3 M). Compared with PEFT method TSAM~\cite{Li_Liu_Wang_Yu_2025}, our Task-Adapter++ also has a comprehensive improvement (from 0.9\% to 7.0\%) on SSv2. 
Our Task-Adapter++ has further expanded the average improvement, 4.3\% on SSv2-Small and 3.9\% on SSv2-Full compared to the SOTA method TSAM~\cite{Li_Liu_Wang_Yu_2025}, indicating that our method can better adapt to the temporal complex FSAR tasks with dual adaptation. 

In addition, we report the results on HMDB51, UCF101 and Kinetics in Table~\ref{tab:sota2}. Especially for 5-shot tasks, our Task-Adapter outperforms the existing SOTA methods MA-FSAR~\cite{xing2023multimodal} by 0.9\% on HMDB51, 0.2\% on UCF101 and 0.8\% on Kinetics, respectively. 
Above our Task-Adapter, Task-Adapter++ achieve higher score by 2.9\% on HMDB51, 0.7\% on UCF101 and 1.0\% on Kinetics, respectively. Compared with full fine-tuning method CLIP-FSAR~\cite{wang2023}, our Task-Adapter++ achieve an average improvement of 6.9\% on HMDB51, 1.3\% on UCF101 and 1.8\% on Kinetics. Compared with PEFT method TSAM~\cite{Li_Liu_Wang_Yu_2025}, our method also has an average improvement of 2.6\% on HMDB51, 0.5\% on UCF101 and 0.2\% on Kinetics. These scene-related datasets require strong background understand ability which means our model can handle complex scene knowledge well.

\begin{table}[]
\centering
\caption{Ablation study of Task-adapter. We choose frozen backbone as the baseline under 5-way 1-shot settings on SSv2-Small.}
  \setlength{\abovecaptionskip}{0.2cm} 
  \centering
 \resizebox{0.65\textwidth}{!}{
 \begin{tabular}{c c c c c } 
 \toprule
 Frozen Backbone & AIM & Task-specific Adaptation & Partial Adapting & Acc\\
 \midrule
 \checkmark  & $\times$ & $\times$& $\times$ & 36.0\\
 $\times$  & $\times$ & $\times$ & $\times$ & 47.8 \\
 \checkmark  & \checkmark & $\times$ &$\times$ &  53.3\\

 \checkmark  & \checkmark & $\times$ &\checkmark &  54.3\\
 \checkmark  & \checkmark & \checkmark  &$\times$   & 55.2\\
 \checkmark  & $\times$ & \checkmark &\checkmark &  56.6\\
 \midrule
 \checkmark  & \checkmark & \checkmark  &\checkmark  & \textbf{60.2}\\
 \bottomrule

 \end{tabular}
 }
 \label{tab:components2}
\end{table}
\begin{table}[]
\centering
\caption{Ablation study of Task-adapter++. Task-Adapter is chosen as the baseline under 5-way 1-shot settings on SSv2-Small.}
  \setlength{\abovecaptionskip}{0.2cm} 
  \centering
 \resizebox{0.65\textwidth}{!}{
 \begin{tabular}{c c c c c } 
 \toprule
 Task-Adapter & Three Sub-actions & Order Adaptation & Cross-modal Alignment& Acc\\
 \midrule
 \checkmark & $\times$& $\times$  & $\times$ & 60.2\\
 \checkmark  & \checkmark &$\times$ & $\times$ &  62.0\\
 \checkmark  & $\times$ & $\times$ & \checkmark & 58.5 \\

 \checkmark  & \checkmark & $\times$ & \checkmark &  62.5\\
 \checkmark  & \checkmark  & \checkmark & $\times$ & 62.8\\
  \midrule
 \checkmark  & \checkmark & \checkmark &\checkmark &  \textbf{63.6}\\

 \bottomrule

 \end{tabular}
 }
 \label{tab:components}
\end{table}
\subsection{Ablation study}
\textbf{Effectiveness of each component.}
To demonstrate the effectiveness of each component, we perform detailed ablation experiments. The ablation study of our Task-Adapter is show in Table~\ref{tab:components2} and Task-Adapter++ is shown in Table~\ref{tab:components}. In Table~\ref{tab:components2}, the first row shows the results of a frozen pre-trained model without any learnable adapters, while the second row refers to the performance of full fine-tuning. Comparing the results of the third and fourth rows, we can see the effectiveness of PEFT over traditional full fine-tuning in few-shot action recognition. Note that the ``Partial Adapting'' in the fourth column of Table~\ref{tab:components2} refers to a common strategy which only adds adapters in the last several ViT layers. We can observe from the table that the improvement of introducing task-specific adaptation into AIM is distinct (1.9\%) when inserting adapters into all the layers of the pre-trained model. Additionally, using partial adapting strategy for AIM also brings an improvement of 1\%. Even so, the improvement of introducing task-specific adaptation into AIM is notably enhanced under the partial adapting strategy, increasing from 1.9\% to 5.9\%. 

In Table~\ref{tab:components}, the first row presents the result of Task-Adapter, while the second row shows the performance when applying the three sub-actions for textual augmentation, which yields a 1.8\% improvement. The third row reports the effect of applying cross-modal alignment with fine-grained temporal modeling, leading to a decline of 1.7\%. Notably, performing the fine-grained alignment operation without enhancing semantic features is redundant, as the original Task-Adapter already accomplishes semantic alignment during feature extraction. The result in the fourth row supports our assumption, where the model's performance increases further from 1.8\% to 2.3\%. Incorporating order adaptation improves the performance of the ``Three Sub-actions''. It models the sequential relationships between sub-actions in more detail. This change increases the performance from 1.8\% to 2.5\% as shown in the fifth row. The last row represents the performance of the complete model, revealing the complementarity and effectiveness of each component.

\begin{figure}[h]
\centering
  \includegraphics[width=.9\textwidth]{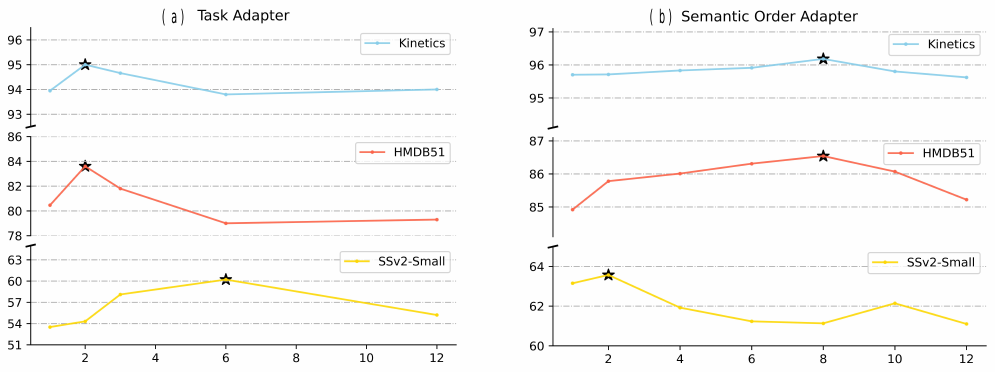}
  \caption{Effect of inserting Task-Adapter into the last $L$ layers (a) and semantic order adapter (b) into the last $M$ layers (e.g., $L,M$ = 1, 2, 3, ..., 12) on 3 different datasets. The best is highlighted with a black star.
  }
\label{fig:layer}
\end{figure}

\textbf{The effect of inserting position of adapters.} 
In this part, we explore the result of different inserting positions. We can see from Fig.~\ref{fig:layer} (a) that inserting our Task-Adapter into the top 2 layers is the best setting for Kinetics and HMDB51. Since these scene-related datasets focus on background understanding, a more lightweight setting with inserting adapters into only the top 2 layers is more appropriate. Inserting the Task-Adapter into the top 6 layers yields the highest performance on SSv2. Since SSv2 requires more intensive temporal modeling than scene-related datasets, a deeper adapter is necessary.

In Fig.~\ref{fig:layer} (b), we demonstrate the impact of the insertion position of the semantic order adapter. Our analysis indicates that for datasets such as SSv2—which already encode rich action information in their original labels (e.g., ``Pouring something into something'')—employing a deep architecture for text sequence modeling is generally unnecessary. Conversely, for datasets with more straightforward action content, such as Kinetics and HMDB51, a more sophisticated modeling approach is warranted. 

In conclusion, for scene-related datasets like UCF101, Kinetics and HMDB51, we choose 2-layer Task-Adapter and 8-layer semantic order adapter. For SSv2, we choose 6-layer Task-Adapter and 2-layer semantic order adapter. A detailed analysis of the tunable parameters will be presented in the following subsection. 

\begin{table}[h]
  \setlength{\abovecaptionskip}{0.2cm} 
  \centering
  \caption{The result that using different modalities of information.}
 \resizebox{0.65\textwidth}{!}{
 \begin{tabular}{l c c c} 
 \toprule
  methods & video matching score  & video-text matching score & both\\
 \midrule
 Task-Adapter  & 53.4 & 50.4 & 60.2\\
 Task-Adapter++  & 53.1 & 54.3 & 63.6 \\
 \bottomrule

 \end{tabular}
 }
 \label{tab:v-s}
\end{table}

\textbf{The result of different modalities of information.} As shown in Table~\ref{tab:v-s}, the integration of multimodal information into the FSAR task significantly enhances the performance of both Task-Adapter and Task-Adapter++. However, we can observe that Task-Adapter achieves an accuracy of only 50.4\% when relying solely on the video-text matching score, suggesting that semantic information is not effectively leveraged. While, Task-Adapter++ can significantly improve the classification accuracy of the semantic branch by 3.9\%. This improvement can be attributed to the proposed order adaptation of semantic information and fine-grained cross-modal matching. Our method effectively addresses the lack of informative cues in video-text matching, balancing the two branches of the model and enhancing their complementary strengths.

\begin{table}[h]
  \setlength{\abovecaptionskip}{0.2cm} 
  \centering
  \caption{The results of different utilization methods for enhanced sub-action description.}
 \resizebox{0.7\textwidth}{!}{
 \begin{tabular}{l | c c c c} 
 \toprule
  methods & w/o sub-actions &average features &concat sub-actions & ours\\ \midrule
  accuracy & 60.2 & 60.7 & 61.4 & \textbf{63.6}\\
 \bottomrule

 \end{tabular}
 }
 \label{tab:river}
\end{table}

\textbf{Different methods for using the sub-actions.} In general, LLMs provides three descriptions of sub-actions. Except for our claimed method in Section~\ref{section3}, it is easy to consider them as a whole. This actually includes two ideas: averaging the three semantic features or concatenating the descriptions together as one prompt. 
Therefore, Table~\ref{tab:river} shows the accuracy with different strategies, i.e., without using sub-actions description, averaging features and concatenating descriptions together. The results not only demonstrate the effectiveness of decomposing actions into sub-actions, but also reveal the information loss caused by averaging and concatenation, thereby validating the rationale behind stage-wise matching

\begin{figure*}[h]
\centering
  \includegraphics[width=.9\textwidth]{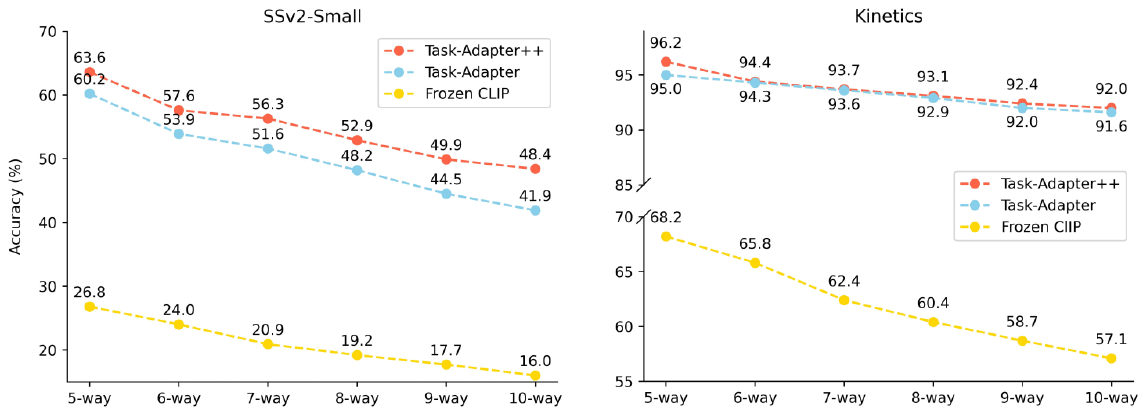}
  \caption{The results of $N$-way 1-shot between different methods on SSv2-Small and Kinetics.
  }
\label{fig:nway}
\end{figure*}

\textbf{$N$-way 1-shot results.} Fig.~\ref{fig:nway} shows that our approach, compared to the original Task-Adapter and the baseline, not only improves absolute accuracy but also better adapts to $N$-way 1-shot learning ($N$ $\ge$ 5), i.e., experiences less relative decline. As $N$ increases, classification difficulty also increases, and model performance decline is inevitable. However, because our approach fully considers both semantic and visual features, its performance is less affected in SSv2-Small. And in Kinetics, the results are nearly saturated. It can be seen that the results descend slightly when $N$ increases, and the gap keeps very small.

\begin{table}[h]
  \setlength{\abovecaptionskip}{0.2cm} 
  \begin{center}
  \caption{The results of different alignment modules on 5-way 1-shot task.}
 \label{tab:align}
 \resizebox{0.5\textwidth}{!}{
 \begin{tabular}{l c | c c } 
\toprule

  \multirow{1}{2cm}{Method} & \multirow{1}{1cm}{setting} & \multirow{1}{2cm}{SSv2-Small} & \multirow{1}{1cm}{Kinetics} \\
  \midrule 
  & 1-shot & 63.6 & 96.2 \\
  1-layer Cross Attnetion & 3-shot & 69.1 & 97.3 \\
  & 5-shot & 73.2 & 97.5 \\
  \midrule
  & 1-shot & 60.4 & 94.3 \\
  2-layer Cross Attnetion & 3-shot & 66.7 & 96.1 \\
  & 5-shot & 71.7 & 97.1 \\
  \midrule
  & 1-shot & 60.6 & 93.3 \\
  3-layer Cross Attnetion & 3-shot & 67.0 & 95.7 \\
  & 5-shot & 71.4 & 96.8 \\
  \midrule
  & 1-shot & 55.4 & 92.3 \\
  Conv1d & 3-shot & 62.9 & 94.9 \\
  & 5-shot & 68.5 & 96.0 \\
\bottomrule
 \end{tabular}
 }
 \end{center}
\end{table}
\textbf{The exploration of different alignment modules.} In Table~\ref{tab:align}, we explore different temporal modeling modules for the alignment task. We utilize multi-layer cross attention and 1D CNN (layer=1, stride=1, kernel=3, channel=512) to align visual features. The results indicate that stacking multi-layer cross attention is ineffective and does not improve results, while 1D CNN shows significant disadvantage over cross attention.

\begin{figure}[h]
\centering
  \includegraphics[width=.5\textwidth]{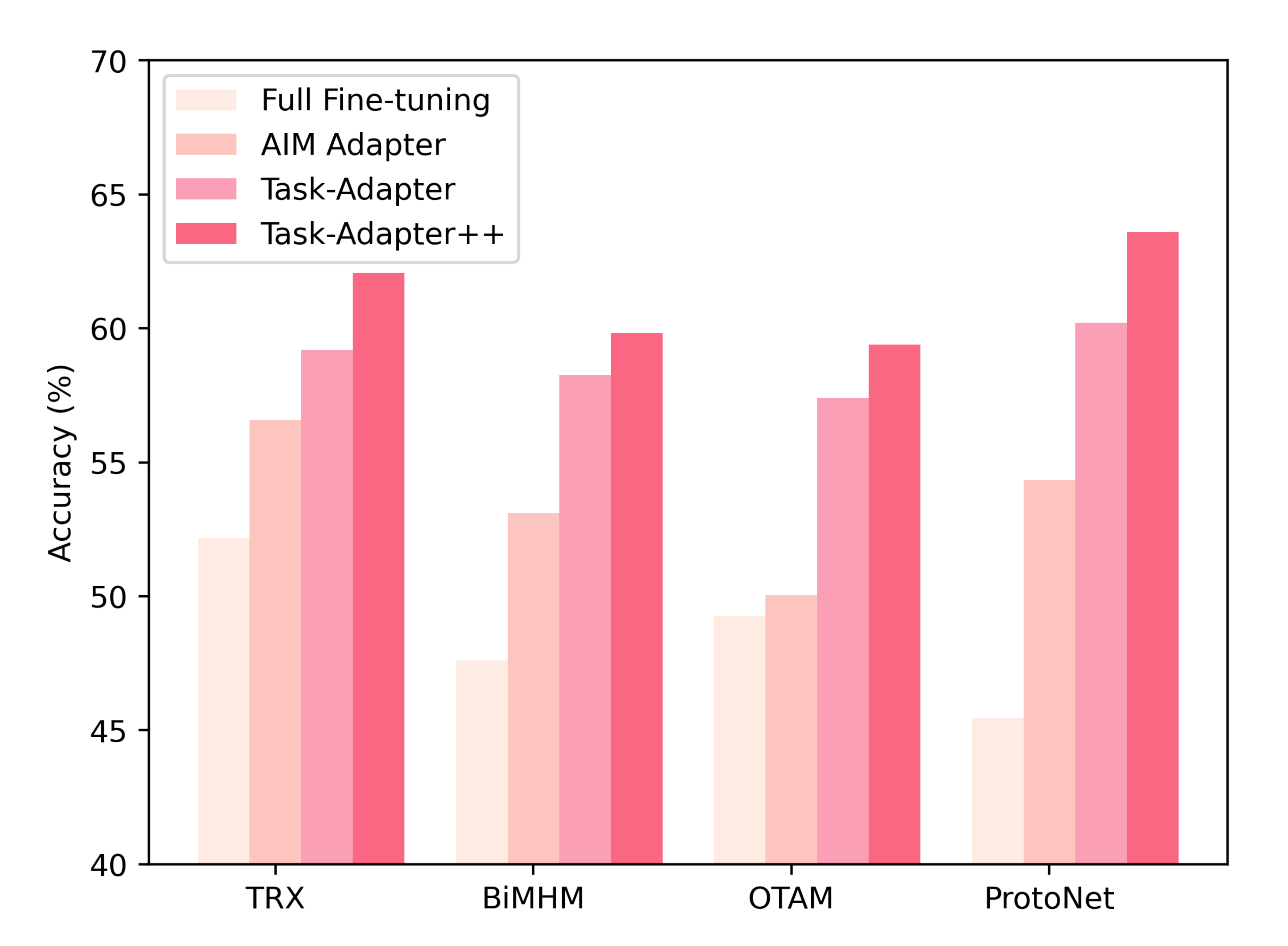}
  \caption{Comparison of the performance achieved by combining different fine-tuning strategies with existing widely used metric measurement methods on SSv2-Small 5-way 1-shot.
  }
\label{fig:pz}
\end{figure}
\textbf{Different fine-tuning strategies using different metric measurement methods.}
We also compare the performance of different fine-tuning strategies using different metric measurement methods to demonstrate the applicability of our Task-Adapter++. Four commonly used methods are taken in this paper, namely TRX~\cite{perrett2021temporal}, Bi-MHM~\cite{wang2022hybrid}, OTAM~\cite{cao2020few} and ProtoNet (e.g., directly averaging the frame-to-frame cosine similarity over temporal dimension without complex metric measurements). Taking full fine-tuning strategy as the baseline, we observe a consistent improvement across all metric measurement methods, as shown in Fig.~\ref{fig:pz}. It should be noted that ProtoNet outperforms all other methods when uniformly using Task-Adapter++. We attribute this to the fact that with our Task-Adapter++, the model significantly improves the utilization of multimodal information and enhances fault tolerance on the basis of Task-Adapter.

\begin{table}[h]
  \setlength{\abovecaptionskip}{0.2cm} 
  \begin{center}
  \caption{The tunable parameter analysis of task adapters and order adapters on SSv2-Small.}
 \label{tab:par}
 \resizebox{0.6\textwidth}{!}{
 \begin{tabular}{l c c | l c c } 
\toprule

 & \multirow{1}{3cm}{Task-Adapter}& & &\multirow{1}{3cm}{Order Adapter} & \\
\hline
  Position & Tunable Param & Acc & Position & Tunable Param & Acc \\
  \midrule 
 All & 14.1 M & 55.2 & All & 1.58 M & 61.1 \\
 Bottom 6 & 7.2 M & 50.7 & Bottom 6 & 790 K & 62.1 \\
 Bottom 2 & 2.4 M & 47.4 & Bottom 2 & 263 K & 62.6 \\
 Top 6 & 7.2 M & 60.2 & Top 6 & 790 K & 61.2 \\
 Top 2 & 2.4 M & 54.3 & Top 2 & 263 K & 63.6 \\
\bottomrule
 \end{tabular}
 }
 \end{center}
\end{table}

\textbf{The tunable parameter analysis.} 
Firstly, we study the accuracy and tunable parameter of different inserting positions of our Task-Adapter in Table~\ref{tab:par} left. On SSv2-Small, inserting our Task-Adapter only into the top 6 layers outperforms inserting it into all layers by a large margin (5\%) with only 7.2 M parameters. We attribute this to the hypothesize that bottom layers focus on generic features that do not require extensive adaptation, while the top layers focus on task-specific and discriminative features that benefit a lot from task-specific adaptation module. 
Additionally, We study the accuracy and tunable parameter of different inserting positions of our semantic order adapter in Table~\ref{tab:par} right. Note that inserting adapters only into the top 2 layers achieves the best performance with only 263 K parameters. Given the small performance gap at the bottom 2/6 layers, we believe that semantic features are not yet well extracted, which is in contrast to the top 2/6 layers where stronger semantic representations are observed. 

In conclusion, for scene-related datasets, Task-Adapter++ utilizes only 2.4 M parameters in the visual branch and 1.1 M parameters in the semantic branch, resulting in a total of \textbf{3.5 M} parameters. In comparison, for temporal-related datasets, the model employs 7.2 M parameters in the visual branch and 263 K parameters in the semantic branch, amounting to \textbf{7.5 M} parameters in total. These two strategies both significantly reduce the number of parameters compared to full fine-tuning, which requires 149.6 M parameters.

\begin{table}[h]
  \setlength{\abovecaptionskip}{0.2cm} 
  \begin{center}
  \caption{Impact of position of Task-MSA and O-MSA inner one ViT layer on SSv2-Small 5-way 1-shot.}
 \label{tab:pos}
 \resizebox{0.6\textwidth}{!}{
 \begin{tabular}{l c c | l c c } 
\toprule

  &\multirow{1}{2cm}{Task-MSA}& &&\multirow{1}{2cm}{O-MSA}  \\
\hline
  Position & & Acc & Position & & Acc \\
  \midrule 
 in front of T\&S-MSA &  & 55.6 & in front of MSA & & 62.2 \\
 between T\&S-MSA&  & 55.5& - & & - \\
 back of T\&S-MSA&  & 60.2 & back of MSA& & 63.6 \\
\bottomrule
 \end{tabular}
 }
 \end{center}
\end{table}

\textbf{Position of the Task-MSA and O-MSA inner ViT layer.} In the visual branch, as illustrated in Section~\ref{section3}, the original AIM comprises two multi-head self-attention (MSA) modules, individually for spatial and temporal modeling. We study the optimal placement of our proposed task-specific MSA (Task-MSA) relative to the existing two MSAs for task-specific modeling. As shown in Table~\ref{tab:pos} left, inserting the Task-MSA after the temporal (T-) and spatial (S-) MSAs achieves the best performance. It is noteworthy that the Task-MSA consistently outperforms the baseline AIM~\cite{yang2023aim}, regardless of its insertion position. These results indicate that the introduction of the Task-MSA indeed enhances few-shot action recognition performance compared to AIM. Furthermore, placing the Task-MSA after the T- and S-MSAs enables maximum extraction of task-specific discriminative features. In the semantic branch, we retain the original Task-Adapter as our image encoder. We find that placing the order MSA (O-MSA) after the original MSA yields higher scores than placing it frontally, as shown in Table~\ref{tab:pos} right. This is probably because modeling the order relationships of sub-action sequences benefits more from refined representations, which are typically captured in later stages of the network. Nevertheless, both configurations outperform the baseline Task-Adapter, demonstrating the effectiveness of the proposed O-MSA
\begin{figure*}[h]
\centering
  \includegraphics[width=1.0\textwidth]{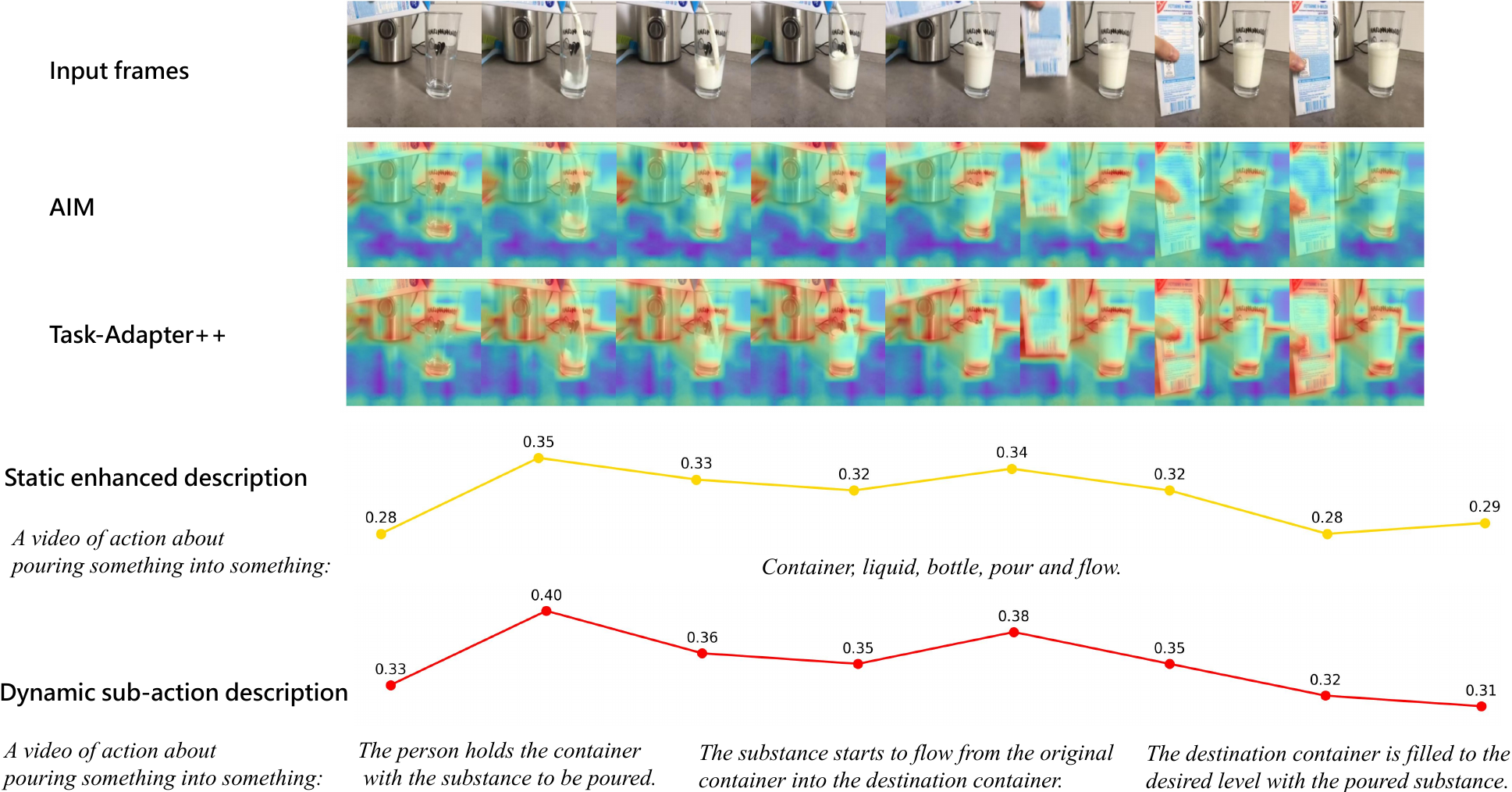}
  \caption{Visualization of the attention maps and cross-modal matching scores for the ``Pouring something into something'' action of different methods. We first present the attention maps generated by AIM and Task-Adapter++ to demonstrate the superiority of our Task-Adapter. Subsequently, we illustrate the cross-modal matching scores at various stages, thereby substantiating the effectiveness of the dynamic sub-action description.}
\label{fig:visualization}
\end{figure*}
\subsection{Visualization results}
To further demonstrate the effectiveness of our proposed Task-Adapter++, we visualize the attention maps and cross-modal matching scores in Fig.~\ref{fig:visualization}. The input frames are sampled from the ``Pouring something into something'' action in SSv2-Small. Notably, the attention map generated by AIM is less effective at capturing informative regions, such as the carton of milk, compared to that produced by our Task-Adapter++. Furthermore, our method employs a dynamic sub-action description that segments the complete action into three distinct stages—beginning, process, and end. In comparison with the static prompt, our dynamic prompting strategy yields significant improvements in fine-grained cross-modal matching performance.

\section{Conclusion}
In this paper, we propose Task-Adapter++ , a novel framework for adapting pre-trained image models to few-shot action recognition. Instead of traditional full fine-tuning, we introduce a parameter-efficient dual adaptation method that exclusively fine-tunes newly inserted adapters, thereby mitigating catastrophic forgetting and overfitting issues. Specifically, we design a task adapter module to enhance discriminative features in the visual domain for given few-shot learning tasks. Concurrently, we generate sub-action descriptions to capture dynamic label information and employ a semantic order adapter to model the sequential structure of sub-actions in the semantic domain. To better align visual and semantic features for cross-modal matching, we further propose a novel fine-grained cross-modal alignment strategy. Experimental results fully demonstrate the effectiveness and superiority of the the proposed Task-Adapter++, which yields significant performance improvements across five few-shot action recognition benchmarks, particularly on the challenging SSv2 dataset.

\section{Acknowledgements}
This work is supported by National Natural Science Foundation of China (No. 62376217, 62301434), Young Elite Scientists Sponsorship Program by CAST (No. 2023QNRC001), and the Joint Research Project for Meteorological Capacity Improvement (Grants 24NLTSZ003).

\bibliographystyle{elsarticle-num-names}
\bibliography{mybibfile}
\newpage

\begin{wrapfigure}{l}{25mm} 
    \includegraphics[width=1in,height=1.25in,clip,keepaspectratio]{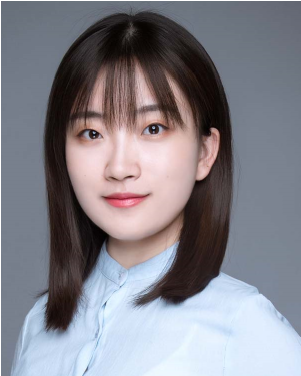}
  \end{wrapfigure}\par
  \textbf{Congqi Cao} (Member, IEEE) received the B.E. degree in information and communication from Zhejiang University, Hangzhou, China, in 2013, and the Ph.D. degree in pattern recognition and intelligent systems from the Institute of Automation, Chinese Academy of Sciences, in 2018. Then, she joined the School of Computer Science, Northwestern Polytechnical University, Xi’an, China, where she is currently an Associate Professor. Her research interests include computer vision, pattern recognition and relative applications, especially on action recognition/anticipation, video anomaly detection/anticipation, and few-shot learning.\par
  \hspace*{\fill}
  \begin{wrapfigure}{l}{25mm} 
    \includegraphics[width=1in,height=1.25in,clip,keepaspectratio]{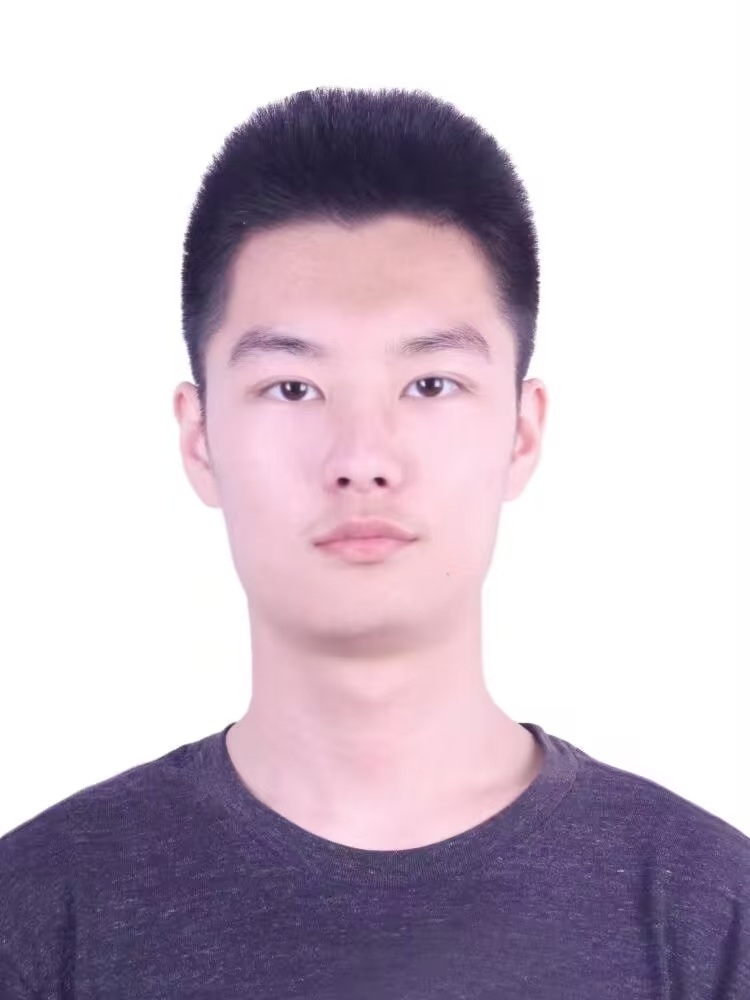}
  \end{wrapfigure}\par
  \textbf{Peiheng Han} received the B.E. degree in artificial intelligence from Northwestern Polytechnical University, Xi’an, China in 2024. He is currently pursuing the M.S. degree in computer technology, School of Computer Science, Northwestern Polytechnical University, Xi’an, China. His research interests include computer vision and few-shot action recognition.\par
\hspace*{\fill}

  \begin{wrapfigure}{l}{25mm} 
    \includegraphics[width=1in,height=1.25in,clip,keepaspectratio]{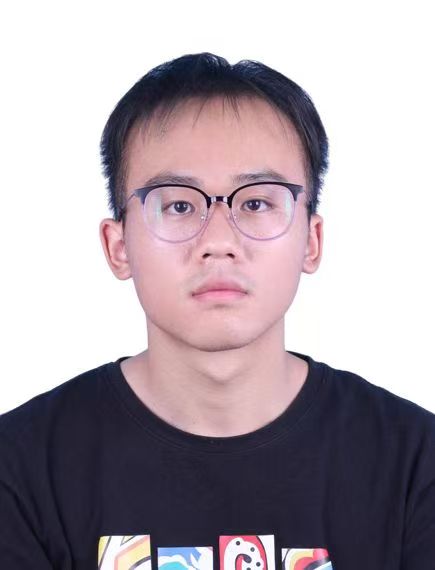}
  \end{wrapfigure}\par
\textbf{ }
  
  \textbf{Yueran Zhang} received the B.E.degree in computer science and technology from Northwestern Polytechnical University, Xi'an, China, in 2022, and the M.E. degree in computer science and technology from Northwestern Polytechnical University, Xi'an, China. His research interests include computer vision, deep learning and relative applications,especially on few-shot action recognition.\par
 \newpage
 \hspace*{\fill}
 \textbf{ }

  \begin{wrapfigure}{l}{25mm} 
    \includegraphics[width=1in,height=1.25in,clip,keepaspectratio]{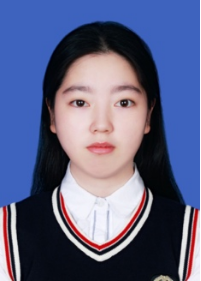}
  \end{wrapfigure}\par
  \textbf{Yating Yu} received the B.E. degree in computer science and technology from Tianjin University of Science and Technology, Tianjin, China, in 2023. She is currently pursuing the Ph.D. degree in computer science and technology with the National Engineering Laboratory for Integrated Aero-Space-Ground-Ocean Big Data Application Technology, School of Computer Science, Northwestern Polytechnical University, Xi’an, China. Her current research interests include video understanding and action recognition in computer vision.\par
  \hspace*{\fill}

\begin{wrapfigure}{l}{25mm} 
\includegraphics[width=1in,height=1.25in,clip,keepaspectratio]{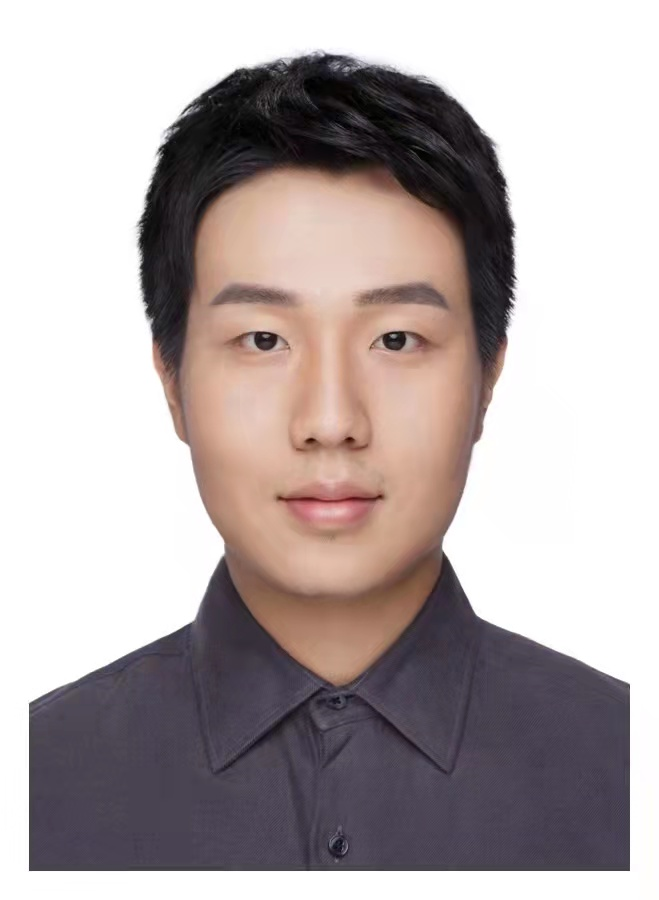}
  \end{wrapfigure}
  \textbf{Qinyi Lv} received the B.S. and Ph.D. degrees from Zhejiang University, Hangzhou, China, in 2013 and 2018, respectively. He is an associate professor with Northwestern Polytechnical University. His recent research interests include RF systems, microwave circuits, radar detection, and inverse scattering problems.\par
\hspace*{\fill}

\subsection*{  }
\begin{wrapfigure}{l}{25mm} 
\includegraphics[width=1in,height=1.25in,clip,keepaspectratio]{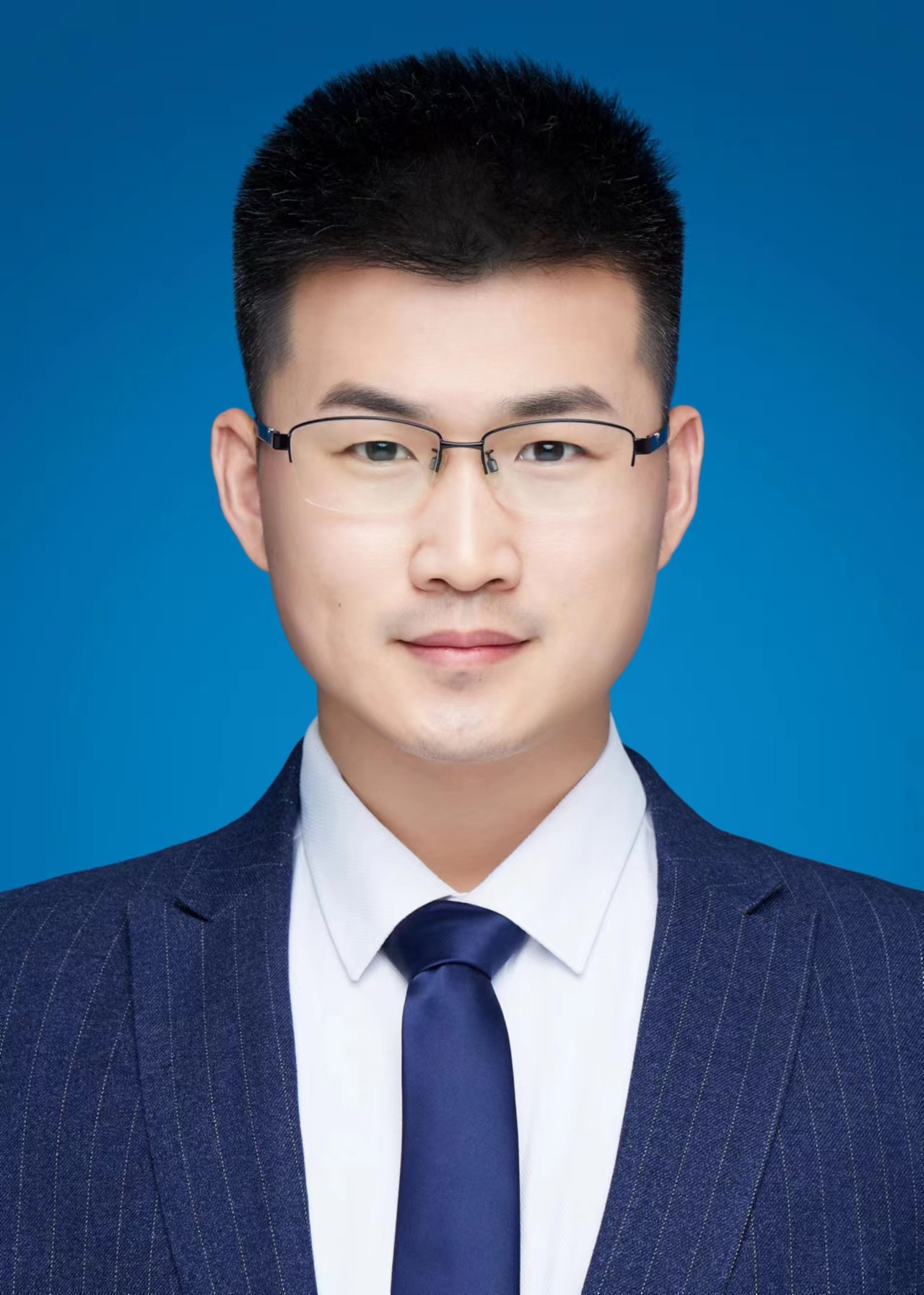}
  \end{wrapfigure}
\textbf{ }

  \textbf{Lingtong Min} 
  received the B.S. degree from Northeastern University, Shenyang, China, in 2012, and the Ph.D. degree from Zhejiang University, Hangzhou, China in 2019. He is an associate professor with Northwestern Polytechnical University. His main research interests are computer vision, pattern recognition, and remote sensing image understanding.
  \par
\hspace*{\fill}
\newpage
      \begin{wrapfigure}{l}{25mm} 
\includegraphics[width=1in,height=1.25in,clip,keepaspectratio]{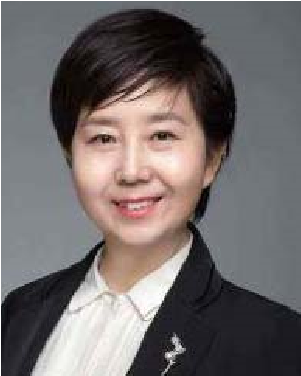}
  \end{wrapfigure}\par
  \textbf{ }
  
  \textbf{Yanning Zhang} (Fellow, IEEE) received the B.E. degree from the Dalian University of Technology, Dalian, China, in 1988, and the Ph.D. degree from the School of Marine Engineering, Northwestern Polytechnical University, Xi’an, China, in 1996. She is currently a Professor with Northwestern Polytechnical University.	Her research interests include computer vision and pattern recognition, image and video processing, and intelligent information processing.
\par
\par
\end{document}